\documentclass{article}

\usepackage[preprint]{icml2026}

\usepackage{microtype}
\usepackage{graphicx}
\usepackage{booktabs}
\usepackage{multirow}
\usepackage{float}
\usepackage{tabularx}
\usepackage{amsmath}
\usepackage{amssymb}
\usepackage{mathtools}
\usepackage{amsthm}
\usepackage{xcolor}
\usepackage[hidelinks]{hyperref}
\usepackage[capitalize,noabbrev]{cleveref}

\hypersetup{
  pdftitle={Balanced State Subspaces with Spectral Rank Allocation for Mamba-2 State Compression},
  pdfauthor={Tingyu Shang, Wenfei Cao},
  pdfsubject={}
}
\makeatletter
\renewcommand{\ICML@preprint}{\textit{Preprint.}}
\makeatother

\newcolumntype{Y}{>{\centering\arraybackslash}X}

\theoremstyle{plain}

\theoremstyle{definition}

\theoremstyle{remark}

\newcommand{\method}{\textsc{BASS}}
\newcommand{\spar}{\textsc{SPAR}}
\newcommand{\sparr}{\textsc{SPAR-R}}
\icmltitlerunning{Balanced State Subspaces with Spectral Rank Allocation}

\begin{document}

\twocolumn[
\icmltitle{Balanced State Subspaces with Spectral Rank Allocation for Mamba-2 State Compression}

\begin{icmlauthorlist}
\icmlauthor{Tingyu Shang}{snu}
\icmlauthor{Wenfei Cao}{snu}
\icmlauthor{Zhi Han}{zky}
\end{icmlauthorlist}

\icmlaffiliation{snu}{School of Mathematics and Statistics, Shaanxi Normal University, Xi'an 710119, China.}
\icmlaffiliation{zky}{State Key Laboratory of Robotics and Intelligent Systems, Shenyang Institute of Automation, 
Chinese Academy of Sciences, Shenyang 110016, China}

\icmlcorrespondingauthor{Wenfei Cao}{caowenf2015@gmail.com}

\icmlkeywords{Mamba-2, state compression, balanced truncation, rank allocation,
long-context retrieval}

\vskip 0.3in
]

\printAffiliationsAndNotice{}

\begin{abstract}
Mamba-2 compresses all context into a fixed-size recurrent state that
dominates decoding memory and bandwidth. Existing compressors quantize state
values or prune coordinates with per-channel scores. Trained states are
nearly rank-one and bias-like, aligned with the mean write; but the output's energy spans an order of magnitude more directions. Per-channel scores see only the
diagonal of the write--read operator, so they cannot select the subspace that
governs the input--output map, which is a rotation of the coordinates. We therefore propose \method{}, which keeps
that subspace by applying balanced truncation exactly via Mamba-2's
scalar-identity transition. Layers differ just as sharply: most are
bias-dominated, while the few that carry long-range associations are the
flattest and need most of their state. We therefore introduce \spar{}, which
redistributes a fixed total rank budget
across layers according to the Hankel spectrum. On Mamba-2 checkpoints from
130M to 2.7B, \method{}+\spar{} halves the state at an average cost of $0.3$
WikiText-2 perplexity points, against $1.6$ for GHOST, while retaining $97\%$
of the dense model's five-digit in-context recall. At a quarter of the state,
giving every layer the same rank keeps $5\%$ of that recall and \spar{}
$77\%$, for a $7\%$ perplexity cost.
\end{abstract}

\section{Introduction}
\label{sec:intro}
\begin{figure*}[t]
\centering
\includegraphics[width=\textwidth]{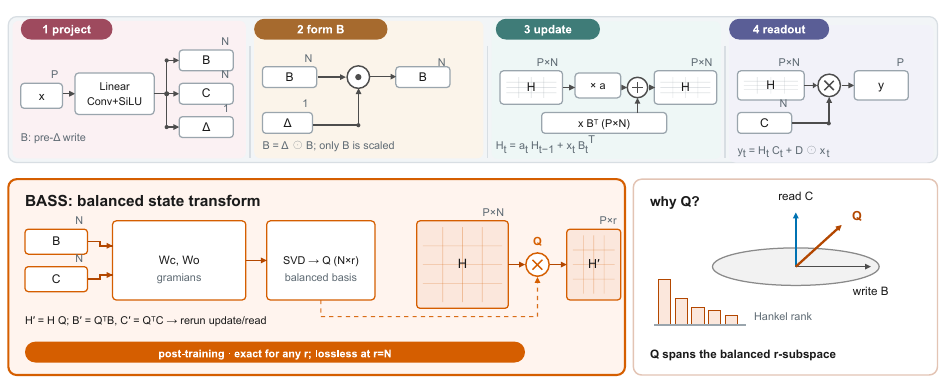}
\caption{\method{} overview. Top: one Mamba-2 layer with pre-$\Delta$ write
$\tilde B_t$, read $C_t$, and step $\Delta_t$, where
$B_t=\Delta_t\odot\tilde B_t$; state and output follow
Eqs.~\eqref{eq:rec}--\eqref{eq:read}. Bottom: \method{} estimates
$W_c,W_o$, keeps the balanced basis $Q$, and applies $H'=HQ$,
$B'=Q^{\top}B_t$, $C'=Q^{\top}C_t$; the compressed layer reruns update and
read with $H',B',C'$. The rotation is exact for any $r$; lossless at $r=N$;
only truncation $r<N$ approximates. Post-training; all other parameters
untouched.}
\label{fig:method}
\end{figure*}

State space models (SSMs) such as Mamba and Mamba-2 have become a
competitive alternative to Transformers for long-sequence modeling: they
scale linearly in sequence length, avoid the growing key--value cache, and
match or exceed attention-based models at small and medium scale
\citep{gu2023mamba,dao2024transformers}. Their efficiency at inference stems from a single mechanism: the entire
context is compressed into a fixed-size recurrent state. For Mamba-2, each layer keeps one matrix state
per head ($N{=}128$ state modes), and generation updates this state in
place at every token, so the state becomes the bottleneck: at batch size
128, the recurrent states of Mamba2-2.7B account for roughly 80\% of
decoding memory \citep{qmamba2025}, and every generated token must stream
them from memory \citep{avmp2026}.

Reducing the cost of the state has therefore become an active goal.
Post-training quantization compresses state values (e.g., Quamba and
Q-Mamba \citep{quamba2024,qmamba2025}). An orthogonal direction prunes
state \emph{coordinates}: GHOST selects state channels by an output-aware
controllability$\times$observability score \citep{ghost2026}, and
\citet{la_state_reduction2026} discard key/query channels in linear
attention. All of these methods
operate in the \emph{native coordinate system} of the state, scoring
coordinates once from calibration statistics; even output-aware
criteria score each coordinate independently, and none can
represent the joint structure that governs how the state as a whole is
written and read. Analyses of the state itself are equally partial:
anisotropy for Mamba-1 \citep{lost_in_state_space2026}, low rank and head
stratification for linear attention
\citep{rank_dynamics2026,la_state_reduction2026}, and output-aware
importance for Mamba-2 \citep{ghost2026}, but never jointly under one
measurement protocol.

In this paper we ask two questions. First, how should we read trained
Mamba-2 states? Second, given that structure, what is the \emph{right}
way to compress them? We \emph{re-examine} Mamba-2 checkpoints from 130M to
2.7B with a single capture pipeline and find two facts about the state's
geometry. (i) State energy is almost rank-one: most heads carry
their energy in at most two directions, and the leading direction recurs
across sequences. (ii) That direction is a bias-like accumulator aligned
with the mean write, while the output's energy is spread over an order of
magnitude more directions; no ordering of native coordinates captures the
subspace that governs the input--output map. Compression should therefore
keep the write--read subspace rather than the coordinates. We turn this
basis into \method{}, a post-training transform that estimates empirical
controllability and observability Gramians, takes their balanced-truncation
subspace, and applies it exactly through Mamba-2's scalar-identity
transition, so the state shrinks from $N$ to $r$ modes while every other
parameter stays untouched (Section~\ref{sec:method});
Figure~\ref{fig:method} summarizes this step.

A third fact concerns the layers rather than the directions. Heterogeneity
is across \emph{layers}, not heads: most layers are effectively rank one or
two and dominated by a running mean, while the few that carry long-range
associations are the spectrally flattest and need most of their state.
Compression should therefore spend the budget where the spectrum says the
state is needed. We turn this into \spar{}, which redistributes a fixed total rank
across layers according to the Hankel spectrum, whose tail is largest
exactly at the layers that carry content
(Section~\ref{sec:method-alloc}).

Experiments on Mamba-2 checkpoints from 130M to 2.7B show that
\method{}+\spar{} halves the state dimension at an average cost of $0.3$
WikiText-2 perplexity points, against $1.6$ for GHOST, while retaining $97\%$
of the dense model's five-digit in-context recall; at a quarter of the state,
giving every layer the same rank keeps only $5\%$ of that recall, while
\spar{} keeps $77\%$ for a $7\%$ perplexity cost; a $2\%$ perplexity
difference can separate allocations whose recall differs sixfold. Random
and anti-allocated controls (ranks permuted across layers) confirm that
where the budget is placed, not merely how much is spent, drives quality. Calibration takes a single forward
pass, an order of magnitude cheaper than GHOST's released protocol
(Section~\ref{sec:experiments}).

Our contributions are:
\begin{itemize}
\item We show that trained Mamba-2 states (130M--2.7B) are almost rank-one
in energy and bias-like, that the subspace carrying the input--output map is
a rotation rather than a set of coordinates, and that the layers which carry
long-range recall are the ones whose spectra are flattest; these observations
motivate the subspace transform and the per-layer budget
(Sections~\ref{sec:analysis} and~\ref{sec:exp-placement}).
\item We propose \method{}, a post-training transform that performs balanced
truncation of the recurrent state using full (non-diagonal) controllability
and observability Gramians and applies the rotation exactly via Mamba-2's
scalar-identity transition, without retraining; to our knowledge, this is the
first balanced-truncation method to use full empirical Gramians for the
recurrent state of a trained selective SSM (Section~\ref{sec:method}).
\item We propose \spar{}, which redistributes a fixed total rank budget
across layers from the Hankel spectrum; at 25\% retention it retains
43--85\% five-digit in-context recall where uniform allocation falls to
0--22\%, while also lowering perplexity. An optional floor (\sparr{})
targets short-range next-word retrieval (Section~\ref{sec:experiments}).
\end{itemize}

\section{Related Work}
\label{sec:related}

\paragraph{State space models.}
Structured SSMs originate with S4 \citep{gu2022s4} and were made
input-dependent (``selective'') by Mamba \citep{gu2023mamba}. Mamba-2
introduced state space duality (SSD): a selective SSM with
scalar-times-identity transition is equivalent to a masked linear
attention and admits tensor-core-friendly computation
\citep{dao2024transformers}. The same duality underlies linear attention
\citep{katharopoulos2020}, whose states share the structure studied here.
Mamba-3 \citep{lahoti2026mamba3} improves state tracking and retrieval
with a trapezoidal, complex-valued rotational update and matches Mamba-2
perplexity at half the state size; this is an architectural reduction,
whereas we study the complementary post-training question of compressing
a trained checkpoint.

\paragraph{State compression.}
Quantization compresses the \emph{values} of the state: Quamba caches the
state in 8 bits with hardware-aware scan kernels \citep{quamba2024};
Q-Mamba handles outliers along the state and channel dimensions to
quantize state caches to 4 bits \citep{qmamba2025}; Quamba2 extends this
to several weight/activation formats and both Mamba generations
\citep{quamba2_2025}, and MambaQuant equalizes channel variances to make
the family 8-bit friendly \citep{mambaquant2025}. These methods are
orthogonal to ours and can be applied after our rotation. Structured
pruning removes \emph{coordinates}: GHOST, the closest prior work,
scores each state channel by
the product of empirical controllability and observability from forward
passes---the diagonal of the Gramians---and prunes the lowest-scoring
channels \citep{ghost2026}. The scalar-identity transition makes the
state's \emph{dynamics} coordinate-wise, but the write and read
couplings are not, so its per-channel score ignores the joint structure
that governs the input--output map (Section~\ref{sec:analysis}).
PerfMamba prunes states
whose gate activity is consistently low \citep{perfmamba2025}, and
Mamba-Shedder removes blocks and sub-modules \citep{mamba_shedder2025}.
In linear attention, \citet{la_state_reduction2026} select key/query
channels with a rank-revealing QR. These methods all operate in the
native coordinate system. Prior analyses report anisotropy and low-rank dynamics in
trained states \citep{lost_in_state_space2026,rank_dynamics2026} and
study what the state encodes
\citep{hidden_attention2025,mamba_knockout2025}; none measures the joint
write--read geometry or uses it for compression.

\paragraph{System-theoretic reduction.}
Balanced truncation \citep{moore1981} retains the subspace that is
simultaneously reachable and observable, ranked by Hankel singular
values. Prior work applies it to linear time-invariant SSMs in three
settings: during training \citep{compressm2026}, to S4's diagonal layers
\citep{ezoe2024}, and for order reduction of deep SSMs
\citep{forgione2024}; Hankel spectra also set layer ranks
\citep{hankelreg2025}, and a recent instrument measures per-input mode
usage in Mamba-1 \citep{exact_instrument2026}. Mamba-2 is time-varying,
so these guarantees do not apply directly, and prior work on trained
Mamba-2 states still selects among native coordinates by diagonal scores
\citep{ghost2026}. We instead estimate full empirical Gramians from
calibration statistics; in this regime the scalar-identity transition
lets us apply the transform exactly, without altering the recurrence.

\section{Preliminaries}
\label{sec:prelim}

\paragraph{Mamba-2 recurrence.}
A Mamba-2 layer applies, per head, the recurrence
\begin{align}
H_t &= a_t\,H_{t-1} + x_t\,B_t^{\top}, \label{eq:rec}\\
y_t &= H_t\,C_t + D \odot x_t, \label{eq:read}
\end{align}
where $H_t \in \mathbb{R}^{P \times N}$ is the recurrent state,
$x_t \in \mathbb{R}^{P}$ (the value/input stream),
$B_t, C_t \in \mathbb{R}^{N}$ are input-dependent write and read
couplings, $D \in \mathbb{R}^{P}$ is a skip connection, and
$a_t \in (0,1)$ is a scalar decay shared by all $N \times P$ state
elements of the head \citep{dao2024transformers}. In the implementation,
$B_t = \Delta_t \odot \tilde B_t$ (with $\tilde B_t$ the pre-$\Delta$ write)
and $C_t$ are produced by linear projections
of the token representation followed by a short depthwise convolution and
a SiLU activation. Throughout, $B$ and $C$ are shared across the heads of
a group (a grouped-query-attention-like structure).
The state is the only context memory of the layer: at sequence length
$L$, the layer stores $P\cdot N$ numbers per head regardless of $L$, and
autoregressive decoding touches all of them at every step.

\paragraph{Balanced truncation.}
For a stable linear time-invariant (LTI) system
$\mathbf{x}_{t+1} = A\mathbf{x}_t + Bu_t$, $y_t = C\mathbf{x}_t$, the
controllability and observability Gramians
$P_c = \sum_{k \ge 0} A^k BB^{\top}(A^{\top})^k$ and
$P_o = \sum_{k \ge 0} (A^{\top})^k C^{\top}C A^k$
quantify how strongly inputs drive each state direction and how strongly
each direction affects outputs. The eigenvalues of $P_cP_o$ are the squared
Hankel singular values $\sigma_i^2 = \lambda_i(P_cP_o)$; balanced truncation
retains the subspace spanned by the leading generalized directions and is
provably near-optimal for LTI systems \citep{moore1981}. For a selective
SSM, $A$ is time-varying and no such guarantee holds; we therefore work
with \emph{empirical} Gramians $W_c,W_o$ (the analogues of $P_c,P_o$)
estimated from calibration trajectories,
and evaluate the resulting transform end-to-end
(Sections~\ref{sec:method} and~\ref{sec:experiments}).
A crucial simplification of Mamba-2 is that $A_t = a_t I$ is a scalar
multiple of the identity: any invertible change of state coordinates
$H \mapsto M H$ leaves the recurrence invariant up to transformed input and
output couplings, because scalar matrices commute with $M$. We exploit this
property to apply the truncated transform exactly.

\section{The Structure of Trained Mamba-2 States}
\label{sec:analysis}

We capture per-layer, per-head states of Mamba2-130M, 370M, 780M, 1.3B and
2.7B~\citep{dao2024transformers} on four WikiText-2 training sequences
\citep{merity2016wikitext}, using windows held out from calibration
(Appendix~\ref{app:capture}). Because the fused implementation does not
expose per-step states, we recompute $x_t, B_t, C_t, \Delta_t$ from the
exact reference equations and verify the replay against the model's
cached recurrence; the maximum relative error is $2.2\times10^{-6}$
across probed layers. We write energy for the mean squared state norm
and anisotropy for the mean pairwise cosine between cross-sequence
states.

\paragraph{Finding 1: energy is almost rank-one.}
Consistent with prior reports of anisotropy and low
rank~\citep{lost_in_state_space2026,rank_dynamics2026},
Table~\ref{tab:state-stats} shows that
per-head spectra are dominated by one direction: the median effective
rank (participation ratio of the squared singular values) is 1.16--1.33,
89.1--92.7\% of heads have effective rank at most two, and the leading
direction carries 87--93\% of the energy. States are strongly
anisotropic: the mean cross-sequence cosine is 0.31--0.43 and reaches
0.99 in early layers.

\begin{table}[t]
\centering
\small
\setlength{\tabcolsep}{4pt}
\begin{tabular}{@{}lcccccc@{}}
\toprule
Model & $L$ & $H$ & erank med.\ (p90) & $\leq 2$ & top-1 & aniso \\
\midrule
130M & 24 & 24 & 1.16 (2.10) & 89.6\% & 0.93 & 0.43 \\
370M & 48 & 32 & 1.17 (1.76) & 91.9\% & 0.92 & 0.39 \\
780M & 48 & 48 & 1.17 (1.76) & 92.7\% & 0.93 & 0.38 \\
1.3B & 48 & 64 & 1.28 (1.93) & 90.7\% & 0.89 & 0.32 \\
2.7B & 64 & 80 & 1.33 (2.07) & 89.1\% & 0.87 & 0.31 \\
\bottomrule
\end{tabular}
\caption{State statistics per model. $L$: layers; $H$: heads. ``erank'':
median effective rank of a head; ``$\leq 2$'': fraction of heads with
effective rank $\leq 2$; ``top-1'': median energy share of the leading
direction; ``aniso'': median per-layer mean pairwise cosine of
cross-sequence states.}
\label{tab:state-stats}
\end{table}

\paragraph{Finding 2: functional content is concentrated, but not in the coordinates.}
The scalar-identity transition makes the channel recurrences
independent, but it does not make the useful subspace coordinate-wise.
The leading energy direction is bias-like: it aligns with the layer's
mean write direction $B$ (median $|\cos|$ 0.88--0.95), whereas
individual writes vary (consecutive $B_t$ cosine 0.51--0.54).
Content-carrying directions therefore have small energy but non-trivial
readout coupling. Keeping four randomly chosen coordinates of 128
retains at most 8.5\% of the output energy $\|C_t^{\top}H_t\|^2$ (median
${\approx}3\%$, against the $4/128{=}3.1\%$ baseline), so channel
independence does not make the state coordinate-sparse. Nor does a
coordinate ordering: even the best $r$ native coordinates---an oracle
over coordinate subsets---capture only 0.69--0.72 of the squared row
mass of the retained balanced subspace at $r{=}64$ (random subspace:
0.55; Appendix~\ref{app:stability}), and at matched rank the balanced
subspace retains 96.6\% of the empirical Hankel energy versus 70.0\% for
the strongest diagonal ordering on 130M at $r{=}32$
(Table~\ref{tab:hankel-retention}). The relevant subspace is defined
jointly by how the state is written and how it is read---the
balanced-truncation criterion we adopt next.

\paragraph{Finding 3: the useful rank varies across layers, not heads.}
State-reduction methods fix one retention ratio for all layers and,
where heads are treated separately, for all heads. Heads within a layer
are similar (median within-layer standard deviation of effective rank
0.16--0.23), so per-head budgets have little room to improve; across
layers, a few early-to-middle layers host heads with effective rank up
to 12.8 while the rest are effectively rank-one, more than an order of
magnitude apart. A uniform budget therefore over-provisions
near-rank-one layers. \spar{} reallocates the same average budget across
layers and improves perplexity; adding the retrieval floor (\sparr{})
recovers retrieval (Section~\ref{sec:exp-alloc})---this is what
per-layer rather than per-head budgets mean.

\section{Method}
\label{sec:method}

We present two components. \method{} is a post-training transform that
estimates empirical Gramians of the state (Section~\ref{sec:method-gram})
and applies the balanced subspace exactly
(Section~\ref{sec:method-transform}); Figure~\ref{fig:method} summarizes it
and fixes notation. \spar{} then allocates ranks across layers on top of the
per-layer spectra (Section~\ref{sec:method-alloc}).

\paragraph{Overview.}
The top strip of Figure~\ref{fig:method} follows one Mamba-2 layer: the input
$x_t$ is projected to the pre-$\Delta$ write $\tilde B_t$, the read $C_t$, and
the step $\Delta_t$; the write is scaled,
$B_t=\Delta_t\odot\tilde B_t$, and the state and output follow
Eqs.~\eqref{eq:rec}--\eqref{eq:read}. \method{} changes only the state and
the couplings $B_t,C_t$; the projections, convolution, $\Delta_t$, $A_t$,
$D$, and normalization are untouched. The lower panel isolates the surgery:
estimate the empirical Gramians $W_c,W_o$ from $B_t,C_t$
(Section~\ref{sec:method-gram}), take the balanced basis
$Q=\mathrm{orth}(L_cV_{:,1:r})\in\mathbb{R}^{N\times r}$
(Section~\ref{sec:method-transform}), and apply
$H'=HQ$, $B'=Q^{\top}B_t$, $C'=Q^{\top}C_t$; the compressed layer then
reruns update and read with $H',B',C'$. The rotation is exact for any $r$;
at $r=N$ it is lossless, and only truncation $r<N$ approximates. \spar{} and
\sparr{} (Section~\ref{sec:method-alloc}) select $r$
per layer and are not shown.

\subsection{Calibrating the state Gramians}
\label{sec:method-gram}

Balanced truncation is defined by the controllability and observability
Gramians of the state; for a trained selective SSM neither has a closed
form, so we estimate both from calibration statistics. For each layer
(and each group, since $B$ and $C$ are shared across heads of a group) we
run the pretrained model over four 1{,}024-token calibration sequences
and collect the write and read couplings $B_t, C_t \in \mathbb{R}^{N}$
after the convolution and activation, together with the states $H_t$ for
diagnostics. We then form the empirical Gramians
\begin{equation}
W_c = \frac{1}{T}\sum_{t=1}^{T} B_t B_t^{\top}, \qquad
W_o = \frac{1}{T}\sum_{t=1}^{T} C_t C_t^{\top},
\label{eq:gramians}
\end{equation}
which measure, respectively, how strongly the input history writes into
each state direction and how strongly each direction is read out. For an
LTI system with $A = aI$, the infinite-horizon Gramians reduce to these
time-averaged covariances up to the common decay weighting; for the
selective case we treat \eqref{eq:gramians} as an empirical analogue
rather than an exact infinite-horizon Gramian. Both are $N \times N$ and
estimated once per layer (4{,}096 tokens,
Appendix~\ref{app:capture}); the cost is a single forward pass over the
calibration set.

\subsection{Balanced subspace and exact transform}
\label{sec:method-transform}

Balanced truncation keeps the $r$-dimensional subspace that is
simultaneously easy to reach (controllable) and easy to observe
(observable). At a fixed $r$ it minimizes the discarded Hankel energy;
directions that are written but never read---or read but never
written---receive small Hankel values and are truncated first. Let
$W_c = L_c L_c^{\top}$ and $W_o = L_o L_o^{\top}$ be the Gramian
factorizations. The balanced coordinates are obtained from the singular
value decomposition
\begin{equation}
L_o^{\top} L_c = U \Sigma V^{\top},
\label{eq:svd}
\end{equation}
and the leading $r$ directions in the original state space are spanned by
$L_c V_{:,1:r}$; we orthonormalize them into $Q \in \mathbb{R}^{N \times r}$
supported on the subspace that is \emph{both} strongly written
and strongly read, ranked by Hankel singular values $\sigma_i$.
Coordinate selection is the special case $Q=I$: it can only keep native
axes, while $Q$ may mix them.

Because Mamba-2 uses a scalar-times-identity transition (Section~\ref{sec:prelim}),
the transform $H \to H Q$ commutes with the recurrence: applying the
rotation to the couplings,
\begin{equation}
B'_t = Q^{\top} B_t, \qquad C'_t = Q^{\top} C_t,
\label{eq:surgery}
\end{equation}
yields a \emph{smaller} selective SSM whose state is
$H Q \in \mathbb{R}^{P \times r}$ and whose outputs coincide with the
original model restricted to the retained subspace. The remaining
parameters---$\Delta$, $A$, $D$, normalization, and the input projections---
are untouched. At $r=N$ with $Q$ orthonormal, the transform is exact; we
verify this at full rank in Section~\ref{sec:experiments}.

In practice, the couplings are produced \emph{after} a depthwise
convolution and a SiLU activation, so the rotation cannot be folded into
the input projection weights exactly. \method{} therefore applies a small
$N \times r$ linear map to $B_t$ and $C_t$ at runtime. This costs
$2Nr$ multiply--accumulates per token per layer---negligible compared with
the $2NrP$ state update itself---while reducing the state memory and the
per-step state traffic by a factor $N/r$. Folding the rotation into the
projections via a short recovery fine-tune, or into the scan kernel, is a
systems extension we leave to future work.

\subsection{SPAR: spectral per-layer allocation of rank}
\label{sec:method-alloc}

The balanced construction provides, for each layer $\ell$, Hankel singular
values $\sigma_1^{(\ell)}\ge\cdots\ge\sigma_N^{(\ell)}$; truncating that
layer to rank $r$ discards the tail energy
$e_\ell(r)=\sum_{i>r}(\sigma_i^{(\ell)})^2$. \spar{} (Spectral Per-layer
Allocation of Rank) turns this into a per-layer budget: given a target
average rank $\bar r$ and a floor $r_{\min}$, it allocates ranks by solving
\begin{equation}
\begin{aligned}
\min_{r_1,\dots,r_L}\quad & \sum_{\ell=1}^{L} e_\ell(r_\ell) \\
\text{s.t.}\quad & \sum_{\ell=1}^{L} r_\ell = L\bar r,
\qquad r_{\min}\le r_\ell\le N .
\end{aligned}
\label{eq:alloc}
\end{equation}
The marginal reductions $e_\ell(r)-e_\ell(r+1)$ are non-increasing in $r$,
so a greedy procedure is optimal for the discrete problem: starting from
$r_\ell=r_{\min}$, repeatedly grant one mode to the layer with the largest
marginal tail reduction until the budget $L\bar r$ is met. A priority
queue implements this in $O(LN\log L)$ time.

\paragraph{SPAR-R: retrieval-protected allocation.}
\label{sec:method-harp}

The Hankel tail that \spar{} minimises is a likelihood-oriented proxy and
does not guarantee retrieval performance (Section~\ref{sec:exp-alloc}).
\sparr{} adds a retrieval-aware correction on top of the same machinery by
measuring retrieval sensitivity with a leave-one-layer-out probe. For each
layer $\ell$, it compresses only that layer to a probe rank
$r_{\mathrm p}$ while keeping every other layer at $\bar r$, evaluates the
model on a held-out retrieval set $\mathcal{D}_{\mathrm{probe}}$ disjoint
from the evaluation set, and records the accuracy drop
\begin{equation}
s_\ell = a(\bar r) - a_\ell ,
\label{eq:sensitivity}
\end{equation}
where $a(\bar r)$ is the accuracy of the uniformly allocated model and
$a_\ell$ that of the probed model. The probe requires $L$ forward passes
and no gradients.

On top of the \spar{} allocation, \sparr{} imposes a \emph{retrieval floor}:
the $k$ layers with the
largest $s_\ell$, ranked by \eqref{eq:sensitivity}, keep
$r_\ell=\bar r$, and \eqref{eq:alloc} is re-solved over the remaining
$L-k$ layers with the remaining budget $(L-k)\bar r$, or with $r_\ell$
capped at $\bar r$ if the floor alone exhausts it. The floor changes only
the per-layer ranks fed to the transform of
Section~\ref{sec:method-transform}; at a matched average rank it adds no
parameters or FLOPs beyond \method{}. A sensitivity-weighted alternative
multiplies each layer's marginal Hankel gain in the greedy step by
\(1+s_\ell/\bar s\), where \(\bar s\) is the mean probe sensitivity; we
evaluate it in Section~\ref{sec:exp-alloc} and Appendix~\ref{app:retr}.
The full procedure costs one calibration pass, one \(N\times N\) singular
value decomposition per layer, \(O(LN\log L)\) allocation, and \(L\) probe
passes; inference adds only the \(2Nr\) map of
Section~\ref{sec:method-transform}.

\section{Experiments}
\label{sec:experiments}

\renewcommand{\topfraction}{0.9}
\renewcommand{\textfraction}{0.1}
\renewcommand{\floatpagefraction}{0.8}
\setcounter{dbltopnumber}{3}

\begin{table*}[t]
\centering
\scriptsize
\setlength{\tabcolsep}{1.3pt}
\begin{tabularx}{\dimexpr\linewidth-1pt\relax}{@{}>{\raggedright\arraybackslash}p{2.7cm} *{10}{Y} >{\centering\arraybackslash}p{2.5cm} >{\centering\arraybackslash}p{2.5cm}@{}}
\toprule
& \multicolumn{10}{c}{WikiText-2 perplexity $\downarrow$} & \multicolumn{2}{c}{Avg.\ increase vs.\ dense} \\
\cmidrule(lr){2-11}\cmidrule(lr){12-13}
& \multicolumn{2}{c}{130M} & \multicolumn{2}{c}{370M} & \multicolumn{2}{c}{780M} & \multicolumn{2}{c}{1.3B} & \multicolumn{2}{c}{2.7B} & \multicolumn{2}{c}{} \\
\cmidrule(lr){2-3}\cmidrule(lr){4-5}\cmidrule(lr){6-7}\cmidrule(lr){8-9}\cmidrule(lr){10-11}
Method & $r{=}64$ & $r{=}32$ & $r{=}64$ & $r{=}32$ & $r{=}64$ & $r{=}32$ & $r{=}64$ & $r{=}32$ & $r{=}64$ & $r{=}32$ & 50\% & 25\% \\
\midrule
Dense & \multicolumn{2}{c}{25.21} & \multicolumn{2}{c}{17.97} & \multicolumn{2}{c}{14.78} & \multicolumn{2}{c}{12.39} & \multicolumn{2}{c}{10.93} & ref. & ref. \\
\addlinespace[2pt]
GHOST & 28.35 & 38.67 & 19.69 & 459.6 & 16.17 & 101.2 & 13.47 & 22.75 & 11.65 & 14.56 & $+1.6$ (9.3\%) & $+111$ (643\%) \\
Magnitude & 1972 & 105.8 & 56.81 & 2771 & 26.77 & 91.71 & 21.72 & 33.49 & 16.03 & 24.17 & $+402$ (1628\%) & $+589$ (3290\%) \\
\addlinespace[2pt]
Saliency & 28.30 & 37.11 & 19.54 & 62.65 & 15.96 & 43.91 & 13.49 & 19.78 & 11.48 & 13.71 & $+1.5$ (8.6\%) & $+19.2$ (116\%) \\
BASS (uniform) & 27.02 & 31.61 & 19.00 & 21.62 & 15.04 & 16.11 & 13.01 & 15.20 & 11.08 & 11.65 & $+0.8$ (4.2\%) & $+3.0$ (16.8\%) \\
BASS + SPAR & $\mathbf{25.91}$ & $\mathbf{28.14}$ & 18.53 & $\mathbf{19.40}$ & $\mathbf{14.86}$ & $\mathbf{15.39}$ & 12.64 & $\mathbf{13.30}$ & 11.05 & 11.32 & $+0.3$ (1.9\%) & $\mathbf{+1.3}$ (6.9\%) \\
BASS + SPAR-R & 25.95 & 30.20 & $\mathbf{18.46}$ & 19.67 & 14.86 & 15.44 & $\mathbf{12.63}$ & 13.41 & $\mathbf{11.03}$ & $\mathbf{11.31}$ & $\mathbf{+0.3}$ (1.8\%) & $+1.7$ (9.1\%) \\
\bottomrule
\end{tabularx}
\caption{WikiText-2 perplexity at $r{=}64/32$ (50\%/25\% retention).
Last two columns: average increase vs.\ dense in ppl points, with relative
change in parentheses (mean over five checkpoints). Bold marks the best
compressed method in each column; Dense is a reference and is not compared.
Official rows use released implementations; ours use \method{}.}
\label{tab:main}
\end{table*}

\begin{table*}[t]
\centering
\scriptsize
\setlength{\tabcolsep}{1.3pt}
\begin{tabularx}{\dimexpr\linewidth-1pt\relax}{@{}>{\raggedright\arraybackslash}p{2.7cm} *{10}{Y} >{\centering\arraybackslash}p{2.5cm} >{\centering\arraybackslash}p{2.5cm}@{}}
\toprule
& \multicolumn{10}{c}{LAMBADA last-word accuracy $\uparrow$} & \multicolumn{2}{c}{Avg.\ change vs.\ dense} \\
\cmidrule(lr){2-11}\cmidrule(lr){12-13}
& \multicolumn{2}{c}{130M} & \multicolumn{2}{c}{370M} & \multicolumn{2}{c}{780M} & \multicolumn{2}{c}{1.3B} & \multicolumn{2}{c}{2.7B} & \multicolumn{2}{c}{} \\
\cmidrule(lr){2-3}\cmidrule(lr){4-5}\cmidrule(lr){6-7}\cmidrule(lr){8-9}\cmidrule(lr){10-11}
Method & $r{=}64$ & $r{=}32$ & $r{=}64$ & $r{=}32$ & $r{=}64$ & $r{=}32$ & $r{=}64$ & $r{=}32$ & $r{=}64$ & $r{=}32$ & 50\% & 25\% \\
\midrule
Dense & \multicolumn{2}{c}{0.449} & \multicolumn{2}{c}{0.557} & \multicolumn{2}{c}{0.607} & \multicolumn{2}{c}{0.648} & \multicolumn{2}{c}{0.703} & ref. & ref. \\
\addlinespace[2pt]
GHOST & 0.355 & 0.193 & $\mathbf{0.463}$ & 0.246 & 0.471 & 0.307 & 0.520 & 0.332 & 0.609 & 0.459 & $-10.9$ ($-18.7\%$) & $-28.6$ ($-49.2\%$) \\
Magnitude & 0.092 & 0.037 & 0.102 & 0.006 & 0.164 & 0.027 & 0.154 & 0.039 & 0.295 & 0.088 & $-43.2$ ($-73.7\%$) & $-55.4$ ($-93.5\%$) \\
Saliency & 0.330 & 0.139 & 0.449 & 0.279 & 0.461 & 0.299 & 0.520 & 0.318 & 0.596 & 0.443 & $-12.2$ ($-21.0\%$) & $-29.7$ ($-51.5\%$) \\
\addlinespace[2pt]
BASS (uniform) & 0.371 & 0.256 & 0.412 & 0.309 & 0.535 & 0.369 & 0.572 & 0.402 & 0.639 & 0.459 & $-8.7$ ($-15.2\%$) & $-23.4$ ($-39.9\%$) \\
BASS + SPAR & 0.320 & 0.221 & 0.428 & 0.369 & 0.602 & 0.410 & 0.576 & 0.369 & 0.596 & 0.356 & $-8.8$ ($-15.8\%$) & $-24.8$ ($-41.9\%$) \\
BASS + SPAR-R & $\mathbf{0.404}$ & $\mathbf{0.352}$ & 0.438 & $\mathbf{0.400}$ & $\mathbf{0.602}$ & $\mathbf{0.504}$ & $\mathbf{0.592}$ & $\mathbf{0.545}$ & $\mathbf{0.664}$ & $\mathbf{0.602}$ & $\mathbf{-5.1}$ ($-9.0\%$) & $\mathbf{-11.2}$ ($-19.5\%$) \\
\bottomrule
\end{tabularx}
\caption{LAMBADA last-word accuracy at the budgets of
Table~\ref{tab:main} (512 examples). Last two columns: average change vs.\
dense in accuracy points, with relative change in parentheses (unweighted
mean over the five checkpoints). Official rows use released
implementations; ours use \method{}. SPAR-R selects layers on held-out
examples (512--767), disjoint from evaluation. Bold marks the best
compressed method in each column; Dense is a reference and is not compared.
Magnitude is a failure floor.}
\label{tab:lambada}
\end{table*}

\begin{table*}[t]
\centering
\scriptsize
\setlength{\tabcolsep}{1.3pt}
\begin{tabularx}{\dimexpr\linewidth-1pt\relax}{@{}>{\raggedright\arraybackslash}p{2.7cm} *{10}{Y} >{\centering\arraybackslash}p{2.5cm} >{\centering\arraybackslash}p{2.5cm}@{}}
\toprule
& \multicolumn{10}{c}{Five-digit passkey retrieval $\uparrow$} & \multicolumn{2}{c}{Avg.\ change vs.\ dense} \\
\cmidrule(lr){2-11}\cmidrule(lr){12-13}
& \multicolumn{2}{c}{130M} & \multicolumn{2}{c}{370M} & \multicolumn{2}{c}{780M} & \multicolumn{2}{c}{1.3B} & \multicolumn{2}{c}{2.7B} & \multicolumn{2}{c}{} \\
\cmidrule(lr){2-3}\cmidrule(lr){4-5}\cmidrule(lr){6-7}\cmidrule(lr){8-9}\cmidrule(lr){10-11}
Method & $r{=}64$ & $r{=}32$ & $r{=}64$ & $r{=}32$ & $r{=}64$ & $r{=}32$ & $r{=}64$ & $r{=}32$ & $r{=}64$ & $r{=}32$ & 50\% & 25\% \\
\midrule
Dense & \multicolumn{2}{c}{0.760} & \multicolumn{2}{c}{0.760} & \multicolumn{2}{c}{0.927} & \multicolumn{2}{c}{0.823} & \multicolumn{2}{c}{0.927} & ref. & ref. \\
\addlinespace[2pt]
GHOST & 0.219 & 0.000 & 0.521 & 0.000 & 0.781 & 0.000 & 0.562 & 0.000 & 0.740 & 0.052 & $-27.5$ ($-32.8\%$) & $-82.9$ ($-98.8\%$) \\
\addlinespace[2pt]
BASS (uniform) & 0.260 & 0.000 & 0.479 & 0.010 & 0.708 & 0.000 & 0.635 & 0.000 & 0.823 & 0.219 & $-25.8$ ($-30.8\%$) & $-79.4$ ($-94.5\%$) \\
BASS + SPAR & $\mathbf{0.760}$ & $\mathbf{0.427}$ & $\mathbf{0.688}$ & $\mathbf{0.469}$ & $\mathbf{0.917}$ & $\mathbf{0.854}$ & 0.750 & $\mathbf{0.646}$ & $\mathbf{0.938}$ & $\mathbf{0.833}$ & $\mathbf{-2.9}$ ($\mathbf{-3.5\%}$) & $\mathbf{-19.4}$ ($\mathbf{-23.1\%}$) \\
BASS + SPAR-R & 0.708 & 0.000 & 0.635 & 0.396 & 0.906 & 0.823 & $\mathbf{0.760}$ & 0.406 & 0.917 & 0.708 & $-5.4$ ($-6.5\%$) & $-37.3$ ($-44.4\%$) \\
\bottomrule
\end{tabularx}
\caption{Five-digit passkey retrieval at the budgets of Table~\ref{tab:main}:
exact recovery of a five-digit key planted in a 512-token context
(96 samples, teacher-forced greedy match). Last two columns: average change
vs.\ dense in accuracy points, with relative change in parentheses
(unweighted mean over the five checkpoints). Bold marks the best compressed
method in each column; Dense is a reference and is not compared.}
\label{tab:passkey}
\end{table*}

\begin{figure*}[t]
\centering
\includegraphics[width=\textwidth]{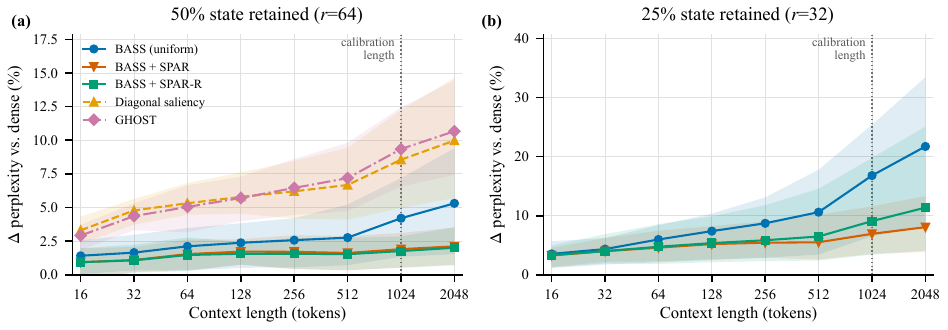}
\caption{Compression cost versus context length. Each line is the mean over
the five checkpoints (130M--2.7B, \(n{=}5\)); shaded bands give the min--max
range across those checkpoints, not a confidence interval, and no significance
test is applied at this \(n\). Calibration runs once at 1024 tokens (dotted
line). (a) 50\% retention. (b) 25\% retention, where GHOST is omitted because
its cost leaves the axis (\(1.4\)--\(25\times\) across models) and is tabulated
in Appendix~\ref{app:length}. Absolute cost falls with length while the
relative cost rises, because dense perplexity itself falls \(9\)--\(11\times\)
over the same range; Appendix~\ref{app:length} reports both views.}
\label{fig:length}
\end{figure*}

\subsection{Setup}
\label{sec:exp-setup}

\paragraph{Models and evaluation.}
We use the public Mamba2-130M, 370M, 780M, 1.3B and 2.7B checkpoints
\citep{dao2024transformers} in float32. Language modeling is evaluated as
chunked perplexity on the WikiText-2 test set (8 segments of 1024 tokens
for 130M/370M/780M and 4 for 1.3B/2.7B; we report the change relative to
the uncompressed model, whose perplexities are 25.21, 17.97, 14.78, 12.39
and 10.93).
Retrieval is evaluated on LAMBADA \citep{paperno2016lambada} with
last-token accuracy over the first 512 examples. All methods, including
the uncompressed baseline, run through the same independent scan
implementation.

\paragraph{Compression methods.}
All methods reduce every layer from $N{=}128$ modes to $r$ modes and are
applied as the transform of \eqref{eq:surgery}:
\emph{saliency} selects the $r$ largest $\mathbb{E}[H^2C^2]$ coordinates,
the diagonal output-aware criterion of GHOST-style channel selection;
\emph{balanced} (\method{}) is the joint-Gramian subspace of
Section~\ref{sec:method-transform}. \method{} + \spar{} applies per-layer
rank allocation, and \method{} + \sparr{} adds the retrieval floor
(Section~\ref{sec:method-harp}).
Energy and random-coordinate selection are tabulated in
Table~\ref{tab:extra-baselines}. Gramians and
saliency are
estimated from four
1024-token sequences drawn from the WikiText-2 training split (the
Gramians use the full sequences; energy and saliency use their
post-warmup tokens), strictly disjoint from all evaluation data; the
sensitivity of \method{} to this calibration set is analyzed in
Appendix~\ref{app:calibration}.

\subsection{Main results}
\label{sec:exp-main}

Table~\ref{tab:main} reports absolute WikiText-2 perplexity; the last two
columns give the relative change, and Figure~\ref{fig:ppl} in the appendix
shows the full rank sweep. BASS+SPAR-R is the best
compressed configuration on perplexity at every scale and budget: at
50\% retention it adds $0.3$ ppl on average versus $1.6$ for GHOST, and
at 25\% retention the gap widens to $1.7$ versus $111$. Even uniform
\method{} costs $+7.2\%$ at 130M versus $+12.5\%$ for GHOST. Three
controls point to the basis: a diagonal-saliency
control tracks GHOST, random selection is far worse, and permuting ranks
is worse than uniform (Appendix~\ref{app:retr}). A full-test-file
evaluation of uniform \method{} reproduces these costs with bootstrap
intervals and makes the 1.3B $r{=}32$ entry conservative
(Appendix~\ref{app:fullppl}).

\paragraph{Official baseline suite.}
We also run the remaining criteria compared in the GHOST paper with the
authors' released implementations: \emph{Magnitude} (static
$\|B_n\|\,\|C_n\|$ scoring) and random selection. Magnitude is
catastrophic and not monotone in budget; Taylor-2 is numerically explosive
at 50\% (Table~\ref{tab:extra-baselines}); random selection sits between
GHOST and Magnitude. All rows are our re-runs under one protocol
(Appendix~\ref{app:baselines}).

\subsection{Where the budget goes}
\label{sec:exp-placement}

Table~\ref{tab:passkey} reports five-digit retrieval. At
50\% retention \spar{} retains $97\%$ of dense recall on average, against
$69\%$ for uniform \method{} and $67\%$ for GHOST; at 25\% it keeps $77\%$
while both fall below $6\%$---a cliff, not a smooth trade in budget.

The cliff is local. A passkey localization probe---one layer at $r{=}8$,
the rest dense---costs $0.67$, $0.45$, and $0.34$ recall at layers 4, 14,
and 10 of the 130M model, and at most $0.05$ elsewhere; a comparably small
set appears at every scale, and three layers out of 24, 48, or 64 carry
the task. It is not the LAMBADA floor probe
(Section~\ref{sec:method-harp}).

Table~\ref{tab:placement} fixes the total budget and varies only its
placement. Giving the sensitive layers a common rank $r_C$ and splitting
the rest evenly raises 130M recall from $0.00$ at $r_C{=}32$ to $0.52$ at
$r_C{=}112$; $r_C{=}96$ reaches $0.427$, matching \spar{} exactly, although
\spar{} splits the same budget as 98/80/69. Permuting \spar{}'s rank
multiset gives at most $0.03$, and a single full-rank layer gives $0.00$:
the effect needs the right set of layers, not merely concentration.
Perplexity is almost flat over the same range ($29.20$ at $r_C{=}64$,
$28.67$ at $r_C{=}112$), so two allocations that differ by $2\%$ in
perplexity differ sixfold in recall.

Why these layers? Their spectra are the flattest: the most sensitive layer
of every model is the flattest or second-flattest, and the sensitive set has
normalized tail $0.10$--$0.40$ against a median of $0.03$--$0.05$.
Alignment is complementary: in 52--79\% of layers the
leading write--read direction $v_1$ has
$|\cos(v_1,\bar B)|{>}0.9$, and the most sensitive layer of every model is
an exception ($0.08$--$0.41$). The two correlate (Spearman
$0.57$--$0.65$) but not uniformly: one sensitive layer per model stays
aligned ($0.76$--$0.98$) yet flat. A budget keyed to
alignment alone would underserve it; the tail does not. In short, the
association lives in the tail of the flattest layers; uniform allocation
cuts that tail, while \spar{} funds it with a median $2.6\times$ the mean
rank and no retrieval signal (Appendix~\ref{app:geometry}).

\subsection{Length robustness}
\label{sec:length}

Because the state is the only channel through which a Mamba-2 layer sees its
past, a compressor that is accurate at the calibration length can still drift
once the recurrence has run longer. Figure~\ref{fig:length} sweeps the
evaluation length from 16 to 2048 tokens with a single calibration at 1024 and
reports the relative cost against the dense model at the same length, averaged
over the five checkpoints.

At 50\% retention every method pays more at long context, but at very different
rates. Between \(L{=}128\) and \(L{=}2048\) the uniform subspace grows by
\(2.2\times\) and the diagonal criteria by \(1.7\)--\(1.9\times\), while
BASS with SPAR-R grows by \(1.3\times\) and ends at \(2.0\%\), \(5.3\times\)
below GHOST's \(10.7\%\). At 25\% retention the separation widens: uniform
reaches \(21.7\%\) at \(L{=}2048\) and SPAR-R stays at \(11.4\%\).

The comparison is controlled for the basis: uniform and SPAR-R share the same
balanced subspace and differ only in the per-layer ranks, so the divergence
isolates the allocation rather than the rotation. GHOST, which also reallocates
a fixed budget but only along native coordinates, tracks the diagonal baseline
instead of closing the gap.

\subsection{Layer-adaptive budgets and retrieval}
\label{sec:exp-alloc}

\paragraph{Allocation versus retrieval.}
Table~\ref{tab:alloc} separates allocation from the retrieval floor on
130M at $\bar r{=}64$: SPAR nearly closes the perplexity gap but loses
LAMBADA accuracy, and the floor restores most of it ($+2.9\%$ at 0.404,
versus $+7.2\%$ at 0.371 for uniform). The sensitivity-weighted variant is
unstable across probe draws, so we use the hard floor; the ordering holds
at every probed scale (Appendix~\ref{app:retr}).

\subsection{Exactness controls}
At $r{=}N$ both the identity and the full-rank balanced transform
reproduce the baseline perplexity exactly ($\Delta = 0.00\%$), and the
offline state recomputation matches the reference implementation to
$2.2\times10^{-6}$ relative error; at $r{<}N$ only the retained subspace changes.

\section{Discussion}
\label{sec:discussion}

\paragraph{Why balanced truncation wins.}
A coordinate-space method is the special case of our transform with
$Q=I$: it can only keep or drop native axes, and its per-channel scores
ignore the cross-channel covariance that joint drops induce. The state's
energy and functional axes differ---a bias-like direction dominates
energy while the readout uses low-energy directions---so any diagonal
criterion trades one against the other; rotating first lets selection
operate in the basis that diagonalizes the input--output map. The
subspace is optimal in the LTI limit, and empirically best or within 0.2
points of the best control at every budget, with write-only, read-only,
and random-rotation controls falling behind as budgets tighten
(Appendix~\ref{app:controls}).

\paragraph{What the budget buys.}
A bias-like state is a running mean: it summarizes the context but cannot
say which key holds which value, and most layers are of this kind, so
truncating them is cheap. Recall lives in the flat layers, in modes beyond
the leading one---which is why a bias-aligned leading direction does not
make a layer cheap (Section~\ref{sec:exp-placement}). Next-token prediction
is a different regime: its LAMBADA-probed bottleneck sits late and stays
bias-dominated, so the spectral allocation and the floor protect different
layers; \sparr{} splits one budget between them.

\paragraph{Interaction with quantization.}
Quantization and subspace reduction are complementary. At $2\times$
compression, 16-bit state quantization is nearly lossless on smaller models;
at $4\times$ and more aggressive compression, low-bit quantization alone
loses accuracy or collapses, making subspace reduction necessary
(Appendix~\ref{app:quant}).

\section{Conclusion}
\label{sec:conclusion}

We studied the recurrent state of Mamba-2 as a compression target and
found that its energy is nearly rank-one, yet its functional content is
concentrated in a subspace that is not aligned with the coordinates.
Based on this, we proposed \method{}, which casts state compression as
balanced truncation of the selective SSM and applies the subspace
transform exactly through the scalar-identity transition, with per-layer
SPAR and a retrieval floor layered on top. At 50\% retention it adds $0.3$
ppl on average across the five checkpoints ($1.6$ for GHOST); on the
five probed scales SPAR-R also recovers $0.02$--$0.07$ LAMBADA accuracy,
and the gap widens at 25\% retention. Future work includes kernel fusion,
extending \method{} to Mamba-1 and complex-state models, and long-context
retrieval.

\section*{Limitations}

\paragraph{Method scope.}
The capacity limit remains: 25\% retention costs $+20$--$+25\%$ on
130M/370M/1.3B and 16 of 128 modes costs $+58\%$ at 130M. Calibration moves
the cost: across disjoint sources the 50\% cost varies by up to 1.5 points,
and 128-token sequences cost up to 1.0/9.9 points at $r{=}64$/32
(Appendix~\ref{app:calibration}).

\paragraph{Evaluation scope.}
We evaluate WikiText-2, LAMBADA, and a 16--2048-token sweep on 130M--2.7B
checkpoints. All checkpoints are trained at 2K tokens, so the sweep measures
within-window robustness, not extrapolation; zero-shot, vision, and hybrid
models remain untested. SPAR-R's floor targets LAMBADA; on five-digit passkey
it trails SPAR at 25\% ($0.00$--$0.82$ vs.\ $0.43$--$0.85$), so
passkey-critical use should prefer SPAR.

\paragraph{Cost and implementation.}
The rotation adds 0.6--0.8\% to prefill latency; calibration is faster and
uses less memory than GHOST (Appendix~\ref{app:cost}). At large batches,
throughput rises $1.5$--$2.2\times$ and peak memory drops $19$--$38\%$;
at batch~1, our unoptimized loop loses 6--10\% (Appendix~\ref{app:decode}).
Quantization uses simulated per-mode integers, not production kernels
(Appendix~\ref{app:quant}).

\section*{Impact Statement}
This paper studies post-training compression of the recurrent state in
Mamba-2. Its intended benefit is lower inference memory and bandwidth for
long-context models. The
main risk is that state compression, like any approximation, can degrade
behavior that aggregate metrics do not capture; we document one such case
(long-range retrieval) in Section~\ref{sec:exp-placement} and provide a
mechanism that preserves it. The method requires only forward passes over
public calibration data, adds no training data, and does not change the
training objective. We see no direct path from this work to harmful
applications beyond those already present in the underlying language
models.

\bibliography{references}

@article{gu2023mamba,
  title = {Mamba: Linear-Time Sequence Modeling with Selective State Spaces},
  author = {Gu, Albert and Dao, Tri},
  journal = {arXiv preprint arXiv:2312.00752},
  year = {2023}
}

@inproceedings{dao2024transformers,
  title = {Transformers are {SSM}s: Generalized Models and Efficient Algorithms Through Structured State Space Duality},
  author = {Dao, Tri and Gu, Albert},
  booktitle = {Proceedings of the 41st International Conference on Machine Learning (ICML)},
  year = {2024}
}

@inproceedings{lahoti2026mamba3,
  title = {Mamba-3: Improved Sequence Modeling using State Space Principles},
  author = {Lahoti, Aakash and Li, Kevin Y. and Chen, Berlin and Wang, Caitlin and Bick, Aviv and Kolter, J. Zico and Dao, Tri and Gu, Albert},
  booktitle = {International Conference on Learning Representations (ICLR)},
  year = {2026}
}

@article{gu2022s4,
  title = {Efficiently Modeling Long Sequences with Structured State Spaces},
  author = {Gu, Albert and Goel, Karan and R{\'e}, Christopher},
  journal = {International Conference on Learning Representations (ICLR)},
  year = {2022}
}

@inproceedings{katharopoulos2020,
  title = {Transformers are {RNN}s: Fast Autoregressive Transformers with Linear Attention},
  author = {Katharopoulos, Angelos and Vyas, Apoorv and Pappas, Nikolaos and Fleuret, Fran{\c{c}}ois},
  booktitle = {Proceedings of the 37th International Conference on Machine Learning (ICML)},
  year = {2020}
}

@article{ghost2026,
  title = {{GHOST}: Unmasking Phantom States in {Mamba2} via Grouped Hidden-state Output-aware Selection and Truncation},
  author = {Menezes, Michael and Kyrillidis, Anastasios},
  journal = {arXiv preprint arXiv:2602.11408},
  year = {2026},
  note = {ICML 2026}
}

@article{perfmamba2025,
  title = {{PerfMamba}: Performance Analysis and Pruning of Selective State Space Models},
  author = {Asif, Ali A. and Kashaniyan, Mahdi and Yu, Simin and Mu{\~n}oz, J. Pablo and Jannesari, Ali},
  journal = {arXiv preprint arXiv:2511.22849},
  year = {2025}
}

@inproceedings{qmamba2025,
  title = {{Q-Mamba}: Towards More Efficient Mamba Models via Post-Training Quantization},
  author = {Chen, Tianqi and Chen, Yuanteng and Xu, Weixiang and Zhu, Zeyu and Wang, Peisong and Cheng, Jian},
  booktitle = {Findings of the Association for Computational Linguistics: ACL 2025},
  year = {2025}
}

@inproceedings{quamba2_2025,
  title = {{Quamba2}: A Robust and Scalable Post-training Quantization Framework for Selective State Space Models},
  author = {Chiang, Hung-Yueh and Chang, Chi-Chih and Frumkin, Natalia and Wu, Kai-Chiang and Abdelfattah, Mohamed S. and Marculescu, Diana},
  booktitle = {Proceedings of the 42nd International Conference on Machine Learning (ICML)},
  year = {2025}
}

@inproceedings{quamba2024,
  title = {{Quamba}: A Post-Training Quantization Recipe for Selective State Space Models},
  author = {Chiang, Hung-Yueh and Chang, Chi-Chih and Frumkin, Natalia and Wu, Kai-Chiang and Marculescu, Diana},
  booktitle = {International Conference on Learning Representations (ICLR)},
  year = {2025}
}

@inproceedings{mambaquant2025,
  title = {{MambaQuant}: Quantizing the Mamba Family with Variance Aligned Rotation Methods},
  author = {Xu, Zukang and others},
  booktitle = {International Conference on Learning Representations (ICLR)},
  year = {2025},
  note = {arXiv:2501.13484}
}

@inproceedings{mamba_shedder2025,
  title = {Mamba-Shedder: Post-Transformer Compression for Efficient Selective Structured State Space Models},
  author = {Mu{\~n}oz, J. Pablo and Yuan, Jinjie and Jain, Nilesh},
  booktitle = {Proceedings of the 2025 Conference of the Nations of the Americas Chapter of the ACL (NAACL)},
  year = {2025}
}

@article{la_state_reduction2026,
  title = {The Key to State Reduction in Linear Attention: A Rank-based Perspective},
  author = {Nazari, Philipp and Rusch, T. Konstantin},
  journal = {arXiv preprint arXiv:2602.04852},
  year = {2026},
  note = {ICLR 2026}
}

@article{rank_dynamics2026,
  title = {State Rank Dynamics in Linear Attention {LLM}s},
  author = {Sun, Ao and Zhang, Hongtao and Zhou, Heng and Ma, Yixuan and Qin, Yiran and Su, Tongrui and Liu, Yan and Ma, Zhanyu and Xu, Jun and Gao, Jiuchong and Hao, Jinghua and He, Renqing},
  journal = {arXiv preprint arXiv:2602.02195},
  year = {2026}
}

@article{exact_instrument2026,
  title = {An Exact Instrument for State Usage in Selective State-Space Models, and the Input-Driven Migration It Reveals},
  author = {Bhattacharya, Raktim},
  journal = {arXiv preprint arXiv:2607.11796},
  year = {2026}
}

@article{lost_in_state_space2026,
  title = {Lost in State Space: Probing Frozen Mamba Representations},
  author = {Wagh, Bhagyashree and Singh, Akash},
  journal = {arXiv preprint arXiv:2605.00253},
  year = {2026}
}

@inproceedings{hidden_attention2025,
  title = {The Hidden Attention of Mamba Models},
  author = {Ali, Ameen and Zimerman, Itamar and Wolf, Lior},
  booktitle = {Proceedings of the 63rd Annual Meeting of the Association for Computational Linguistics (ACL)},
  year = {2025}
}

@inproceedings{mamba_knockout2025,
  title = {Mamba Knockout for Unraveling Factual Information Flow},
  author = {Endy, Nir and Grosbard, Idan Daniel and Ran-Milo, Yuval and Slutzky, Yonatan and Tshuva, Itay and Giryes, Raja},
  booktitle = {Proceedings of the 63rd Annual Meeting of the Association for Computational Linguistics (ACL)},
  pages = {23457--23477},
  address = {Vienna, Austria},
  publisher = {Association for Computational Linguistics},
  year = {2025}
}

@article{avmp2026,
  title = {Asymmetric Virtual Memory Paging for Hybrid Mamba-Transformer Inference},
  author = {Nguyen, An Xuan},
  journal = {arXiv preprint arXiv:2605.22416},
  year = {2026}
}

@article{moore1981,
  title = {Principal Component Analysis in Linear Systems: Controllability, Observability, and Model Reduction},
  author = {Moore, Bruce C.},
  journal = {IEEE Transactions on Automatic Control},
  volume = {26},
  number = {1},
  pages = {17--32},
  year = {1981},
  doi = {10.1109/TAC.1981.1102568}
}

@inproceedings{compressm2026,
  title = {The Curious Case of In-Training Compression of State Space Models},
  author = {Chahine, Makram and Nazari, Philipp and Rus, Daniela and Rusch, T. Konstantin},
  booktitle = {International Conference on Learning Representations (ICLR)},
  year = {2026},
  note = {arXiv:2510.02823}
}

@article{ezoe2024,
  title = {Model Compression Method for {S4} with Diagonal State Space Layers using Balanced Truncation},
  author = {Ezoe, Haruka and Sato, Kazuhiro},
  journal = {arXiv preprint arXiv:2402.15993},
  year = {2024}
}

@article{forgione2024,
  title = {Model Order Reduction of Deep Structured State-Space Models: A System-Theoretic Approach},
  author = {Forgione, Marco and Mejari, Manas and Piga, Dario},
  journal = {arXiv preprint arXiv:2403.14833},
  year = {2024}
}

@inproceedings{hankelreg2025,
  title = {Hankel Singular Value Regularization for Highly Compressible State Space Models},
  author = {Schwerdtner, Paul and Berman, Jules and Peherstorfer, Benjamin},
  booktitle = {Advances in Neural Information Processing Systems (NeurIPS)},
  year = {2025},
  note = {arXiv:2510.22951}
}

@inproceedings{merity2016wikitext,
  title = {Pointer Sentinel Mixture Models},
  author = {Merity, Stephen and Xiong, Caiming and Bradbury, James and Socher, Richard},
  booktitle = {International Conference on Learning Representations (ICLR)},
  year = {2017}
}

@inproceedings{paperno2016lambada,
  title = {The {LAMBADA} Dataset: Word Prediction Requiring a Broad Discourse Context},
  author = {Paperno, Denis and Kruszewski, Germ{\'a}n and Lazaridou, Angeliki and Pham, Quan Ngoc and others},
  booktitle = {Proceedings of the 54th Annual Meeting of the Association for Computational Linguistics (ACL)},
  year = {2016}
}
\bibliographystyle{icml2026}

\clearpage
\appendix
\noindent\textbf{Organization.}
This appendix is organized by the question it answers; the A--N letters
match the main text. A--B cover the capture protocol and a
released-vs-rerun audit of baseline numbers. C, D, I, and J test
sensitivity and attribution: calibration robustness, off-diagonal
coupling and subspace stability, single-sided and random-rotation
controls, and Hankel-energy retention. E, G, and H report layer budgets,
calibration and prefill cost, and decode measurements. K--L close with
full-test-set evaluation and matched-budget quantization; M reports the
rank sweep and N the context-length evaluation. Readers in a hurry
should start with B, I, and K.

\section{State Capture Protocol}
\label{app:capture}

Capturing recurrent states from fused SSM kernels is unreliable: the fused
scan does not expose per-step states, and installing optimized kernels
disables the reference forward path that calibration hooks rely on. We
therefore recompute states offline. For each layer we register forward
hooks on the input projection and the depthwise convolution, reconstruct
$x_t, B_t, C_t, \Delta_t$ with the exact equations of the reference
implementation (including the SiLU activation, the causal padding of the
convolution, the softplus reparameterization of $\Delta$, and the shared
$B, C$ within a group), and replay the recurrence
$H_t = \exp(\Delta_t A) H_{t-1} + (\Delta_t B_t)\, x_t^{\top}$ with the
model's own parameters. We verify the replay against the cached recurrent
state maintained by the reference implementation at several probe layers
and model scales; the maximum relative error is $2.2\times10^{-6}$.

Calibration uses four 1024-token sequences from the WikiText-2 training
split: energy and saliency accumulate over their post-warmup tokens
(256--1023), and the Gramians over the full sequences. The diagnostic
analysis of Section~\ref{sec:analysis} uses the same calibration windows
(tokens 256--512) and held-out windows (tokens 640--896) of four disjoint
training sequences; all perplexity and retrieval evaluation is on the
test split. The surgery experiments reuse the same rank grid for all
methods and evaluate on the same token segments, so all comparisons are
paired.

\section{Baseline Reproduction Details}
\label{app:baselines}

All official rows run the authors' released \texttt{mamba2\_ghost}
implementation with its default soft pruning (\texttt{is\_hard=false}) and
its calibration protocol (128 WikiText-2 samples of 128 tokens, batch 8,
sequential layer-wise pruning); every pruned and uncompressed model is
evaluated with our protocol. Released values are word-level perplexities
from the paper or repository, whereas our reruns are token-level, so
Table~\ref{tab:repro} reports the metric per row rather than a $\Delta$.
We count a
method as reproduced when it lands in the same order of magnitude and
preserves the ordering against the dense model.

\begin{table*}[t]
\centering
\scriptsize
\setlength{\tabcolsep}{3pt}
\begin{tabular}{@{}llllll@{}}
\toprule
Method & Model / budget & Released & Ours & Metric & Reading \\
\midrule
GHOST-2 & 130M / 50\% & $+3.21$ & $+3.14$ & word / token ppl & match \\
Random & 130M / 50\% & 45.6 & 47.5 & token ppl & match \\
Taylor-2 & 370M / 50\% & 12647 & 13851 & word / token ppl & same order \\
Taylor-2 & 1.3B / 50\% & 13.94 & 539 & word / token ppl & released likely no-op \\
Taylor-2 & 130M / 50\% & $1.0{\cdot}10^{16}$ & $4.3{\cdot}10^{18}$ & word / token ppl & both catastrophic \\
Magnitude & 130M / 50\% & 910 & 1972 & word / token ppl & metric gap \\
\bottomrule
\end{tabular}
\caption{Reproduction audit. Released values are word-level perplexities
from the GHOST paper or repository; ours are reruns with the released
compressor under our evaluation. No $\Delta$ is reported because the
metrics differ. ``Reading'' states the interpretation defended in the
text.}
\label{tab:repro}
\end{table*}

\paragraph{Where the two runs diverge.}
Taylor-2 at 1.3B is near-dense in the released sweep, whereas our rerun
reaches 539 ppl. The compressor skips layers for which no calibration
gradient is available (its salience map is never populated), the released
run peaks at 44.9\,GB and uses 1024-token calibration, and all layers
prune 64/128 in our check, so we treat the released run as likely not
pruned. Magnitude's two-fold gap is consistent with word-level versus
token-level scoring. For 370M/25\%, the released pipeline is far worse
than our joint re-implementation of the same diagonal criterion; both
remain behind \method{}, and the residual gap is consistent with
sequential versus joint selection, so we report both.

\paragraph{Robustness of the 25\% collapse.}
The 25\% collapse persists under the authors' calibration protocol, a
32k-token window, and a second calibration draw
(Table~\ref{tab:ghost-robust}). Hard truncation
reproduces the soft results with 31--32 of 128 states retained per layer,
and \method{} stays at $+18$--$26\%$ under the same window.

\begin{table}[t]
\centering
\small
\setlength{\tabcolsep}{3pt}
\begin{tabular}{@{}lccc@{}}
\toprule
Setting & 130M & 370M & 1.3B \\
\midrule
Replica (Table~\ref{tab:main}) & $+53.4$ & $+2458$ & $+83.7$ \\
Replica, 32k window & $+47.3$ & $+1719.7$ & $+106.3$ \\
Official calib., 32k window & $+47.3$ & $+1719.3$ & $+61.7$ \\
Official calib., draw 2, 32k & -- & $+1748.9$ & $+62.5$ \\
Replica calib., draw 2, 32k & -- & $+2063.6$ & -- \\
Hard truncation & $+53.4$ & $+2458$ & $+83.7$ \\
\midrule
\method{} (ours), 32k window & $+25.6$ & $+21.9$ & $+18.1$ \\
\bottomrule
\end{tabular}
\caption{GHOST-2 perplexity change at 25\% under different calibration
protocols, calibration draws and evaluation windows (all relative to the
same uncompressed model). The collapse persists in every setting. Hard
truncation removes channels physically and retains
31--32 of 128 states per layer. ``--'': not run.}
\label{tab:ghost-robust}
\end{table}

\paragraph{Released sweep and additional baselines.}
The released 1.3B sweep is qualitatively identical
(Table~\ref{tab:official-sweep}): GHOST degrades smoothly but is severe
at high sparsity, Taylor-2 jumps between mid and high sparsity, and
Magnitude is unstable from the lowest sparsity.
Table~\ref{tab:extra-baselines} lists Taylor-2
at 780M/2.7B and the energy and random controls under our protocol.

\begin{table}[t]
\centering
\small
\begin{tabular}{lccccc}
\toprule
Sparsity & 0.1 & 0.3 & 0.5 & 0.7 & 0.9 \\
\midrule
Dense & \multicolumn{5}{c}{13.17} \\
GHOST-2 & 13.24 & 13.51 & 14.23 & 16.16 & 25.07 \\
Magnitude & 427.7 & 22.39 & 29.34 & 39.96 & 69.49 \\
Taylor-2 & 13.18 & 13.26 & 13.94 & 4255 & 4690 \\
Random & 13.52 & 15.04 & 17.77 & 30.14 & 64.21 \\
SparseGPT & 13.17 & 13.19 & 13.25 & 13.52 & 15.51 \\
\bottomrule
\end{tabular}
\caption{Word-level WikiText-2 perplexity from the authors' released
sweep on Mamba2-1.3B. SparseGPT is unstructured and does not reduce the
state dimension; it is shown for reference.}
\label{tab:official-sweep}
\end{table}

\begin{table*}[t]
\centering
\scriptsize
\setlength{\tabcolsep}{2pt}
\begin{tabular}{@{}l cccccccccc@{}}
\toprule
& \multicolumn{10}{c}{WikiText-2 perplexity $\downarrow$} \\
\cmidrule(lr){2-11}
& \multicolumn{2}{c}{130M} & \multicolumn{2}{c}{370M} & \multicolumn{2}{c}{780M} & \multicolumn{2}{c}{1.3B} & \multicolumn{2}{c}{2.7B} \\
\cmidrule(lr){2-3}\cmidrule(lr){4-5}\cmidrule(lr){6-7}\cmidrule(lr){8-9}\cmidrule(lr){10-11}
Method & $r{=}64$ & $r{=}32$ & $r{=}64$ & $r{=}32$ & $r{=}64$ & $r{=}32$ & $r{=}64$ & $r{=}32$ & $r{=}64$ & $r{=}32$ \\
\midrule
Taylor-2 (released code, our run) & $4.3{\cdot}10^{18}$ & $4.5{\cdot}10^{34}$ & 13851 & 11654 & $1.7{\cdot}10^{5}$ & $1.2{\cdot}10^{6}$ & 539.0 & 813.4 & 511.7 & 542.9 \\
Energy (ours) & 28.16 & 38.64 & 20.07 & 24.47 & 16.11 & 21.04 & 13.63 & 17.32 & 11.67 & 13.74 \\
Random (ours) & 39.23 & 84.54 & 28.21 & -- & 26.20 & -- & 18.38 & -- & 14.02 & -- \\
\bottomrule
\end{tabular}
\caption{Additional baselines moved from Table~\ref{tab:main} to keep the
main comparison focused. Taylor-2 entries for 780M/2.7B are evaluated in
bf16 (fp32 exceeds 24\,GB); energy selects the $r$ largest
$\mathbb{E}[H^2]$ coordinates, random draws coordinates uniformly, and both
use the BASS protocol and the same evaluation as Table~\ref{tab:main}.
``--'': not run.}
\label{tab:extra-baselines}
\end{table*}

\section{Calibration Sensitivity}
\label{app:calibration}

The balanced basis is estimated from finite calibration statistics, so
its sensitivity to the calibration set matters. Table~\ref{tab:calib}
re-estimates the Gramians from disjoint calibration sources and
re-runs the \method{} surgery at 50\% and 25\%. Across train and
validation calibration sets the 50\% perplexity cost varies by at most 1.5 points;
at 25\% the variation reaches 7.7 points on the 130M model. The final
row of Table~\ref{tab:calib} calibrates on the evaluation text as a
leakage control; all reported results use train calibration.
Ridge shrinkage of the Gramians ($\varepsilon \in
\{0.01,0.05,0.2\}$) and doubling the calibration set did not improve
results further.

Subspace \emph{identity} is not stable at this calibration budget:
pairwise principal angles
between balanced subspaces from different sources reach ${\sim}85^\circ$
in the tail of the Hankel spectrum, and the rank-$r$ Hankel operator
itself changes by $O(1)$ in relative Frobenius norm, because the
eigenvalues beyond the leading one form a near-degenerate cluster.
What is stable is \emph{behavior}: the Hankel spectra agree across
sources (e.g., layer-0 leading eigenvalues $6.21/0.24/0.075$ vs.\
$5.83/0.22/0.072$), and the resulting perplexity is consistent
(Table~\ref{tab:calib}). The comparison to baselines is robust to the
calibration source: even the weakest source at 130M/50\%
($+8.18\%$) remains clearly better than GHOST ($+12.5\%$),
and at 1.3B all sources lie within 0.2 points and are $1.7\times$
better. We therefore report single-source (train) calibration and treat
tail-subspace variability as an inherent property of finite-sample
balanced truncation at this scale.

\begin{table}[H]
\centering
\footnotesize
\setlength{\tabcolsep}{2pt}
\begin{tabular}{@{}lcccc@{}}
\toprule
 & \multicolumn{2}{c}{130M} & \multicolumn{2}{c}{1.3B} \\
 \cmidrule(lr){2-3}\cmidrule(lr){4-5}
Calibration source & $r{=}64$ & $r{=}32$ & $r{=}64$ & $r{=}32$ \\
\midrule
Train, 4 seq.\ (reported) & $+7.18$ & $+25.38$ & $+5.06$ & $+22.75$ \\
Validation, 4 seq. & $+8.18$ & $+33.06$ & $+5.00$ & $+22.34$ \\
Train, 8 seq. & $+7.38$ & $+27.09$ & $+4.89$ & $+22.02$ \\
Eval.\ text (leaky, excluded) & $+3.50$ & $+13.96$ & $+1.02$ & $+8.95$ \\
\bottomrule
\end{tabular}
\caption{WikiText-2 perplexity change of \method{} under different
calibration sources. The last row calibrates on the evaluation text as a
leakage control; all reported results use train calibration.}
\label{tab:calib}
\end{table}

\paragraph{Calibration sequence length.}
The reported protocol uses 1024-token sequences, which estimate the
steady-state write/read statistics; GHOST instead calibrates on
128-token sequences. To separate the protocol from the method, we
re-calibrate \method{} on 128 samples of 128 tokens (batch 8, 16k
tokens; Table~\ref{tab:calib-len}). Shorter sequences cost up to 1.0
point at $r{=}64$ and up to 9.9 points at $r{=}32$ (130M), an effect of
the same order as the calibration-source variation above; the ranking
against GHOST is unchanged---at 128$\times$128, \method{} still costs
$+8.1$/$+6.2$/$+6.0\%$ at 50\% versus $+12.5$/$+9.6$/$+8.7\%$ for
GHOST, and remains far ahead at 25\%.

\begin{table}[H]
\centering
\small
\setlength{\tabcolsep}{4pt}
\begin{tabular}{@{}lcccc@{}}
\toprule
& \multicolumn{2}{c}{$r{=}64$} & \multicolumn{2}{c}{$r{=}32$} \\
\cmidrule(lr){2-3}\cmidrule(lr){4-5}
Model & 128$\times$128 & 4$\times$1024 & 128$\times$128 & 4$\times$1024 \\
\midrule
130M & $+8.14$ & $+7.18$ & $+35.25$ & $+25.38$ \\
370M & $+6.24$ & $+5.73$ & $+23.97$ & $+20.34$ \\
1.3B & $+5.98$ & $+5.06$ & $+25.59$ & $+22.75$ \\
\bottomrule
\end{tabular}
\caption{Calibration-length control: \method{} perplexity change when
the Gramians are estimated from 128-token sequences (GHOST's protocol,
16k tokens) instead of the reported 1024-token sequences.}
\label{tab:calib-len}
\end{table}

\section{Off-Diagonal Coupling and Retained-Subspace Stability}
\label{app:stability}

\paragraph{Coupling structure.}
The empirical Gramians are dominated by their leading direction: a
rank-one matrix with that direction already has an off-diagonal
Frobenius fraction of $0.92$--$0.98$ at $N{=}128$, above our measured
$0.84$--$0.94$, so raw off-diagonal mass reflects the rank-one component
rather than cross-channel structure per se. The functionally relevant
question is whether \emph{native coordinates} can represent the subspace
that the joint write--read geometry selects. They cannot: for the
retained balanced subspace at $r{=}64$, the best 64 native coordinates
capture only $0.69$--$0.72$ of the achievable squared row mass (random
subspace: $0.55$; a coordinate-aligned one: $1.0$; at $r{=}32$:
$0.53$--$0.54$ vs.\ $0.32$)---coordinate selection and joint subspace
selection are genuinely different families, consistent with the control
results above.

\paragraph{Stability of the retained subspace.}
Across disjoint calibration draws (the default 4$\times$1024 training
sequences, the validation split, 16 training sequences, and the leaky
test-text control; up to six pairs per model), the retained top-64
subspace overlaps the default draw in $0.76$--$0.91$ mean squared
principal cosines ($0.67$--$0.88$ at rank 32), against $0.50$/$0.25$ for
random subspaces of the same dimension; the retained Hankel tail mass
agrees to $\le 0.004$ ($\le 0.008$ at rank 32). Individual principal
directions can rotate (maximum per-layer angles up to $90^\circ$) because
the kept spectrum is graded with near-degenerate tail values, so the
exact basis is not identified by finite calibration; the operational
object---the subspace span and the end-to-end cost---is stable across
draws (Table~\ref{tab:calib}).

\section{Layer-Adaptive Budgets and Retrieval}
\label{app:retr}

\paragraph{Allocation ablations.}
Table~\ref{tab:alloc} studies layer-adaptive allocation on the 130M model
at a matched average budget ($\bar r{=}64$). The Hankel tail proxy cuts
the perplexity cost by $2.6\times$ but costs LAMBADA accuracy
(0.371$\to$0.320): it measures likelihood sensitivity, not retrieval.
Per-layer floors do not recover it (0.313 at floor 32); anti-allocating
the same ranks is worse than uniform ($+8.47\%$ vs.\ $+7.18\%$), so the
per-layer distribution matters.

\paragraph{Retrieval probe.}
The probe compresses one layer at a time to $r_{\mathrm p}{=}8$ while all
other layers stay at $\bar r$, and evaluates 256 held-out LAMBADA
examples (indices 512--767, disjoint from the 512 evaluation examples);
Table~\ref{tab:probe} lists the most sensitive layers and probe times. One
late layer dominates in every model, and the ranking correlates only
moderately with SPAR's ranking (e.g., $+0.41$, $+0.41$ and $-0.01$ at
130M, 370M and 1.3B).

\begin{figure}[H]
\centering
\includegraphics[width=\linewidth]{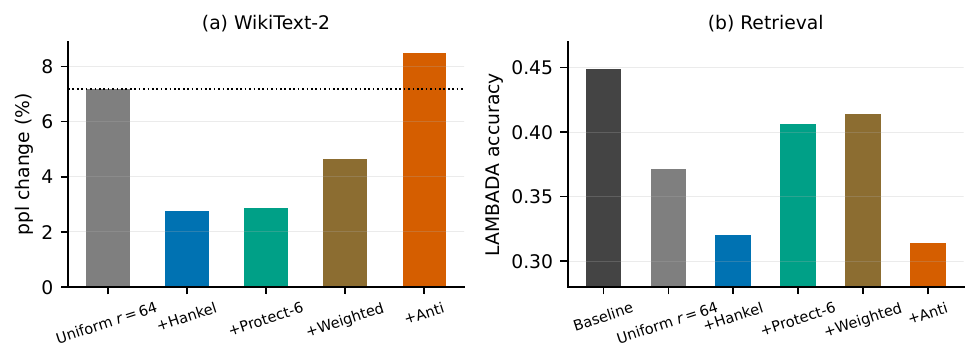}
\caption{Uniform, SPAR, and the two SPAR-R variants (retrieval floor with
$k{=}6$, and sensitivity-weighted), plus an anti-allocated budget, at 50\%
reduction (130M):
the retrieval-aware budgets dominate uniform allocation on both
perplexity and LAMBADA.}
\label{fig:alloc}
\end{figure}

\paragraph{Retrieval floor.}
The floor keeps the six most sensitive layers at $r_\ell=\bar r$ and
re-solves \eqref{eq:alloc} on the remaining layers with the remaining
budget (ties below one point are broken arbitrarily); the
sensitivity-weighted variant is defined in Section~\ref{sec:method-alloc}.
Table~\ref{tab:retr}
compares the two variants with uniform allocation at every scale: the
floor dominates on both axes. The floor budgets are stable across
disjoint probe draws ($\le0.1$ ppl-point and $\le0.02$ LAMBADA
differences), whereas the weighted variant is not: at 1.3B it costs
$+2.45\%$ under the first draw but $+59\%$ under the second, so we
recommend the floor. A paired bootstrap over the 512 evaluation examples
for the floor-minus-uniform difference gives 95\% confidence intervals of
$[+0.008, +0.059]$ at 130M ($p{=}0.016$) and $[+0.004, +0.049]$ at 370M
($p{=}0.041$). On the \emph{full} LAMBADA test set (4{,}897 examples,
excluding the probe window 512--767) the gains are significant at every
scale: $+0.0145$ $[0.007, 0.022]$ ($p{=}2\times10^{-4}$), $+0.0259$
$[0.019, 0.033]$ ($p<10^{-4}$) and $+0.0384$ $[0.031, 0.046]$
($p<10^{-4}$) at 130M/370M/1.3B (the earlier 512-example evaluation
reported the 1.3B margin as within noise, $p{=}0.11$; the larger
evaluation resolves it). Figure~\ref{fig:alloc} shows the per-layer
budgets for the 130M model.

\begin{table}[H]
\centering
\small
\begin{tabular}{lccc}
\toprule
Configuration (130M) & ppl change & LAMBADA \\
\midrule
Uncompressed baseline & $0.00\%$ & 0.449 \\
Uniform $r{=}64$ & $+7.18\%$ & 0.371 \\
Adaptive $\bar r{=}64$, floor 2 & $+2.78\%$ & 0.320 \\
Adaptive $\bar r{=}64$, floor 32 & $+2.81\%$ & 0.313 \\
Retrieval floor ($k{=}6$) & $+2.92\%$ & 0.404 \\
Retrieval floor, weighted & $+3.24\%$ & 0.422 \\
Anti-allocated $r{=}64$ & $+8.47\%$ & 0.314 \\
\midrule
Uniform $r{=}32$ & $+25.38\%$ & 0.256 \\
Adaptive $\bar r{=}32$, floor 2 & $+11.60\%$ & 0.221 \\
\bottomrule
\end{tabular}
\caption{Allocation ablations at a matched average budget on
Mamba2-130M. SPAR improves perplexity but trades off
retrieval; per-layer rank floors do not recover it, while the retrieval
floor does.}
\label{tab:alloc}
\end{table}

\begin{table}[H]
\centering
\footnotesize
\setlength{\tabcolsep}{2pt}
\begin{tabular}{@{}lcccccc@{}}
\toprule
& \multicolumn{2}{c}{Uniform $\bar r{=}64$} & \multicolumn{2}{c}{SPAR-R (floor)} & \multicolumn{2}{c}{SPAR-R (weighted)} \\
\cmidrule(lr){2-3}\cmidrule(lr){4-5}\cmidrule(lr){6-7}
Model & ppl & LAMB. & ppl & LAMB. & ppl & LAMB. \\
\midrule
130M & $+7.18$ & 0.371 & $+2.92$ & 0.404 & $+3.24$ & 0.422 \\
370M & $+5.73$ & 0.412 & $+2.75$ & 0.438 & $+2.05$ & 0.563 \\
780M & $+1.71$ & 0.535 & $+0.52$ & 0.602 & $+1.15$ & 0.600 \\
1.3B & $+5.0$ & 0.572 & $+1.92$ & 0.592 & $+59.0$ & 0.440 \\
2.7B & $+1.34$ & 0.639 & $+0.93$ & 0.664 & -- & -- \\
\bottomrule
\end{tabular}
\caption{Retrieval-aware allocation across scales. The probe selects
layers on held-out LAMBADA examples (512--767), disjoint from the
evaluation set (0--511). The retrieval floor ($k{=}6$) protects the six
most probe-sensitive layers; the weighted variant is unstable across probe
draws. The 2.7B weighted row was not run: the fp32 baseline needs more free
memory than the shared GPU slice (marked ``--'').}
\label{tab:retr}
\end{table}

\begin{table}[H]
\centering
\small
\setlength{\tabcolsep}{3pt}
\begin{tabular}{@{}llc@{}}
\toprule
Model & probe-sensitive layers (drop, pts) & probe time \\
\midrule
130M & 20 (16.4), 19 (7.4), 10 (2.3), 17 (2.3) & 218\,s \\
370M & 38 (15.6), 32 (3.5), 35 (2.7), 28 (2.3) & 779\,s \\
780M & 33 (15.6), 37 (13.7), 9 (4.3), 30 (2.7) & 996\,s \\
1.3B & 33 (26.2), 6 (1.6), 25 (1.2), 37 (0.8) & 1362\,s \\
2.7B & 48 (18.4), 49 (8.2), 29 (5.5), 30 (3.9) & 3079\,s \\
\bottomrule
\end{tabular}
\caption{Leave-one-layer-out retrieval probe ($r_{\mathrm p}{=}8$ for the
probed layer, $\bar r{=}64$ otherwise; held-out LAMBADA examples
512--767, disjoint from the evaluation set).}
\label{tab:probe}
\end{table}

\begin{table}[H]
\centering
\scriptsize
\setlength{\tabcolsep}{3pt}
\begin{tabular}{@{}l ccccc c@{}}
\toprule
Allocation & 130M & 370M & 780M & 1.3B & 2.7B & 130M ppl \\
\midrule
Uniform ($r_\ell{=}32$) & 0.000 & 0.010 & 0.000 & 0.000 & 0.219 & 31.61 \\
Concentrate $r_C{=}48$ & 0.000 & 0.198 & 0.167 & 0.156 & 0.438 & 29.99 \\
Concentrate $r_C{=}64$ & 0.083 & 0.375 & 0.594 & 0.458 & 0.583 & 29.20 \\
Concentrate $r_C{=}80$ & 0.260 & 0.396 & 0.771 & 0.615 & 0.656 & 28.69 \\
Concentrate $r_C{=}96$ & 0.427 & 0.448 & 0.823 & 0.677 & 0.740 & 28.61 \\
Concentrate $r_C{=}112$ & 0.521 & 0.479 & 0.865 & 0.667 & 0.750 & 28.67 \\
SPAR & 0.427 & 0.469 & 0.854 & 0.646 & 0.833 & 28.14 \\
SPAR, permuted & 0.000 & 0.010 & 0.031 & 0.010 & 0.021 & 29.9--38.4 \\
\bottomrule
\end{tabular}
\caption{Placement ablation at 25\% retention (five-digit passkey recall).
Concentrate $r_C$ gives the three layers identified for each model by
single-layer compression a common rank $r_C$ and splits the remaining budget
evenly over the other layers, so the total rank is unchanged. SPAR uses its
own allocation (98/80/69 at 130M); permuted shuffles that rank multiset
across layers (maximum over three seeds).}
\label{tab:placement}
\end{table}

\section{State Geometry}
\label{app:geometry}

Table~\ref{tab:geometry} reports the per-layer statistics behind
Section~\ref{sec:exp-placement}. $v_1$ is the leading balanced direction
$L_cV_{:,1}$ normalized; $\bar B$ is the layer-mean write; tail(32) is the
normalized Hankel tail beyond rank 32. Spearman correlation between
tail(32) and $1-|\cos(v_1,\bar B)|$ is $0.57$--$0.65$ across the five
models.

\begin{table*}[t]
\centering
\footnotesize
\setlength{\tabcolsep}{5pt}
\begin{tabular}{@{}lccccccc@{}}
\toprule
Model & $L$ & $|\cos|>0.9$ & median tail(32) & sensitive layers & tail(32) & $|\cos|$ & tail rank \\
\midrule
130M & 24 & 79\% & 0.034 & 4, 10, 14 & 0.36, 0.28, 0.21 & 0.41, 0.03, 0.98 & 0, 2, 3 \\
370M & 48 & 71\% & 0.045 & 15, 17, 19 & 0.33, 0.24, 0.30 & 0.38, 0.92, 0.34 & 0, 3, 1 \\
780M & 48 & 71\% & 0.039 & 9, 14, 31 & 0.34, 0.37, 0.10 & 0.13, 0.32, 0.86 & 1, 0, 7 \\
1.3B & 48 & 69\% & 0.040 & 6, 20, 23 & 0.23, 0.31, 0.14 & 0.65, 0.18, 0.98 & 4, 0, 6 \\
2.7B & 64 & 52\% & 0.044 & 8, 9, 21 & 0.35, 0.40, 0.29 & 0.03, 0.08, 0.76 & 1, 0, 3 \\
\bottomrule
\end{tabular}
\caption{Layer geometry behind Section~\ref{sec:exp-placement}.
$|\cos|$ is $|\cos(v_1,\bar B)|$; tail rank is counted from the flattest
layer. The sensitive layers and their three values are in increasing layer
order.}
\label{tab:geometry}
\end{table*}

\section{Computational Cost}
\label{app:cost}

Table~\ref{tab:calib-cost} compares the one-time calibration cost of
\method{} and GHOST, measured on one
RTX\,3090 in float32, under each method's released protocol and under a
token-matched swap. \method{} calibrates from four 1024-token sequences
with a single forward pass and a per-layer $128\times128$
eigendecomposition; GHOST's released protocol uses 128 samples of 128
tokens (batch 8, sequential layer-wise pruning). Under the released
protocols \method{} is $8.8$--$10.3\times$ faster and peaks lower; at
matched token budgets it is still $3.2$--$3.3\times$ faster at 16k
tokens and $7.7$--$9.0\times$ faster at 4k tokens, where GHOST also
exceeds the 24\,GB card at 1.3B. GHOST's cost is dominated by its
sequential layer-wise processing and is nearly token-independent
(13.1\,s at 4k versus 14.9\,s at 16k on 130M).

\begin{table}[H]
\centering
\footnotesize
\setlength{\tabcolsep}{2pt}
\begin{tabular}{@{}lcccc@{}}
\toprule
& BASS & BASS & GHOST & GHOST \\
& 4$\times$1024 & 128$\times$128 & 128$\times$128 & 4$\times$1024 \\
\midrule
130M & 1.7 / 2.3 & 4.6 / 2.3 & 14.9 / 3.3 & 13.1 / 10.2 \\
370M & 3.7 / 4.0 & 11.3 / 4.0 & 37.2 / 5.7 & 33.3 / 15.1 \\
780M & 5.2 / 6.8 & 17.0 / 6.9 & 53.7 / 10.6 & -- \\
1.3B & 6.8 / 10.2 & 22.0 / 10.3 & 70.4 / 16.3 & OOM \\
\bottomrule
\end{tabular}
\caption{One-time calibration cost (median seconds / peak GB,
RTX\,3090, float32) under each method's released protocol and under a
token-matched swap. GHOST's 4$\times$1024 run exceeds 24\,GB at 1.3B; the
780M GHOST 4$\times$1024 entry was not run because the shared card had less
than 16\,GB free.}
\label{tab:calib-cost}
\end{table}

Table~\ref{tab:state-cost} lists the recurrent-state size, and
Table~\ref{tab:prefill-cost} the cost of a 1024-token prefill (batch 1,
float32), for the uncompressed model and \method{} at $r{=}64$ and
$r{=}32$. All rows run the same chunked scan implementation, so the
comparison isolates the retained subspace: applying an identity
$N\times N$ map to the uncompressed state costs only 0.6--0.8\% (the
``+map'' column), so the runtime rotation is negligible. The compressed
model is $1.4\times$ ($r{=}64$) to $1.8\times$ ($r{=}32$) faster than
the uncompressed scan and peaks lower in memory at 130M and 370M; at
1.3B the uncompressed scan exceeds the 24\,GB card while the compressed
variants fit. These are prefill measurements under our reference
implementation; decode-side throughput and memory are measured in
Appendix~\ref{app:decode}.

\begin{table}[H]
\centering
\small
\setlength{\tabcolsep}{3pt}
\begin{tabular}{@{}lccc@{}}
\toprule
Model & $N{=}128$ & $r{=}64$ & $r{=}32$ \\
\midrule
130M & 18.0 & 9.0 & 4.5 \\
370M & 48.0 & 24.0 & 12.0 \\
780M & 72.0 & 36.0 & 18.0 \\
1.3B & 96.0 & 48.0 & 24.0 \\
2.7B & 160.0 & 80.0 & 40.0 \\
\bottomrule
\end{tabular}
\caption{Recurrent-state size summed over layers (MB, float32).}
\label{tab:state-cost}
\end{table}

\begin{table}[H]
\centering
\footnotesize
\setlength{\tabcolsep}{2pt}
\begin{tabular}{@{}lcccc@{}}
\toprule
Model & baseline & +map & $r{=}64$ & $r{=}32$ \\
\midrule
130M & 244.6 / 5.12 & $+0.8\%$ & 173.0 / 4.01 & 137.9 / 3.61 \\
370M & 657.2 / 12.63 & $+0.6\%$ & 460.4 / 10.20 & 367.2 / 9.18 \\
1.3B & OOM & OOM & 968.2 / 22.50 & 782.7 / 20.47 \\
\bottomrule
\end{tabular}
\caption{Prefill at $L{=}1024$, batch 1, float32, one RTX\,3090
(median ms / peak GB over five runs). All rows use the same chunked
scan; ``+map'' applies an identity map to the uncompressed state. At
1.3B the uncompressed scan exceeds 24\,GB. The 780M row is not reported:
the shared GPU slice had less than 16\,GB free; its state sizes are in
Table~\ref{tab:state-cost}.}
\label{tab:prefill-cost}
\end{table}

\section{Decode Measurements}
\label{app:decode}

Table~\ref{tab:decode} measures token-by-token decoding with cached
convolution buffers and recurrent states: the same reference loop (plain
PyTorch, no fused kernels) runs the uncompressed model and \method{} at
$r{=}64$/$32$ with the runtime rotation included, so the comparison
isolates the retained state size. The loop reproduces the prefill logits
to $2.6\times10^{-4}$ max absolute error and the $r{=}N$ transform to
the same tolerance. At batch~1, launch overhead dominates: the 130M model
is 3--7\% faster with the smaller state, while 370M/1.3B are 6--10\%
slower from the per-step rotation, and no memory difference is visible.
At larger batches the state read--write traffic dominates: the 130M model
decodes $1.6$--$2.2\times$ faster and the 1.3B model gains
$1.5$--$1.65\times$; peak memory falls with $N/r$. Absolute rates reflect
the unoptimized loop; production kernels would move the crossover batch
lower.

\begin{table}[H]
\centering
\small
\setlength{\tabcolsep}{3pt}
\begin{tabular}{@{}lcccccc@{}}
\toprule
& & \multicolumn{3}{c}{decode tokens/s} & \multicolumn{2}{c}{peak GB} \\
\cmidrule(lr){3-5} \cmidrule(lr){6-7}
Model & $B$ & dense & $r{=}64$ & $r{=}32$ & dense & $r{=}64$ / $r{=}32$ \\
\midrule
130M & 1 & 66.9 & 69.2 & 71.9 & 2.53 & 2.52 / 2.51 \\
130M & 128 & 4050 & 6555 & 8717 & 6.32 & 4.74 / 3.95 \\
370M & 1 & 37.0 & 34.2 & 34.7 & 7.19 & 7.17 / 7.15 \\
370M & 32 & 1131 & 1052 & 1074 & 10.31 & 9.47 / 9.05 \\
780M & 1 & 37.5 & 35.0 & 35.0 & 2.99 & 2.96 / 2.94 \\
780M & 32 & 873.6 & 1082.7 & 1076.2 & 5.42 & 4.23 / 3.63 \\
1.3B & 1 & 37.7 & 33.8 & 35.2 & 5.24 & 5.19 / 5.17 \\
1.3B & 32 & 664 & 992 & 1094 & 8.55 & 6.92 / 6.11 \\
2.7B$^{\dagger}$ & 1 & 23.9 & 21.9 & 22.0 & 10.25 & 10.17 / 10.13 \\
2.7B$^{\dagger}$ & 8 & 186.6 & 170.5 & 172.0 & 6.39 & 5.74 / 5.41 \\
\bottomrule
\end{tabular}
\caption{Decode throughput and peak GPU memory (one RTX\,3090), measured
over 128 steps after warmup with a cached-state reference loop; rotation
included for \method{} rows. All rows are float32 except the
$^{\dagger}$2.7B rows, which are bf16 because fp32 needed more free memory
than the shared card had; they are not directly comparable to the fp32 rows.}
\label{tab:decode}
\end{table}

\section{Rotation and Single-Sided Controls}
\label{app:controls}

To isolate whether the gain comes from the \emph{balanced} basis rather
than from rotation or from either Gramian alone, we re-run the surgery
with three controls under the identical protocol: (i) \emph{write-only},
the top-$r$ eigenvectors of $W_c$; (ii) \emph{read-only}, the top-$r$
eigenvectors of $W_o$; (iii) \emph{random rotation}, a random orthogonal
$N\times r$ basis with the same retention. Table~\ref{tab:controls}
reports perplexity change. At 25\% retention the joint balanced subspace
is best on all four models; at 50\% it is best or within 0.1 points of
the better single-sided control on 1.3B; and a random orthogonal rotation
with the same retention is catastrophic, so retaining coordinates is not
enough---the basis must be aligned with the state's write--read geometry.

\begin{table}[H]
\centering
\small
\setlength{\tabcolsep}{3pt}
\begin{tabular}{@{}llcccc@{}}
\toprule
Model & $r$ & BASS & write-only & read-only & rand-rot \\
\midrule
130M & 64 & $\mathbf{+7.2}$ & $+7.4$ & $+10.0$ & $+78.7$ \\
130M & 32 & $\mathbf{+25.4}$ & $+28.6$ & $+34.7$ & $\infty^{*}$ \\
370M & 64 & $\mathbf{+5.7}$ & $+6.2$ & $+6.1$ & $+135.4$ \\
370M & 32 & $\mathbf{+20.3}$ & $+22.5$ & $+60.1$ & $+239.9$ \\
780M & 64 & $\mathbf{+1.7}$ & $+2.6$ & $+3.1$ & $1.4{\cdot}10^{5}$ \\
780M & 32 & $\mathbf{+9.0}$ & $+12.6$ & $+12.7$ & $+348.3$ \\
1.3B & 64 & $+5.0$ & $+6.1$ & $+4.9$ & $+31.3$ \\
1.3B & 32 & $\mathbf{+22.7}$ & $+27.7$ & $+23.3$ & $+206.2$ \\
\bottomrule
\end{tabular}
\caption{WikiText-2 perplexity change (\%) under matched surgery with
single-sided Gramian bases and a random orthogonal rotation.
$^{*}$Diverged to $4\times10^{21}$ ppl.}
\label{tab:controls}
\end{table}

\section{Hankel-Energy Retention: Coordinate Orderings vs.\ Balanced Subspace}
\label{app:hankel-retention}

For each layer we compute the fraction of the empirical Hankel spectral
energy $\operatorname{tr}(W_cW_o)$ retained by an $r$-dimensional
subspace. For a native-coordinate subset $K$ this is
$\operatorname{tr}((W_c)_{KK}(W_o)_{KK})/\operatorname{tr}(W_cW_o)$,
whereas the balanced subspace retains
$\sum_{i\le r}\sigma_i^2/\sum_i\sigma_i^2$. Coordinate orderings use the
layer-mean per-coordinate score: $\mathbb{E}[H^2]$ (energy),
$\mathbb{E}[H^2C^2]$ (saliency), and
$\operatorname{diag}(W_c)\odot\operatorname{diag}(W_o)$ (the strongest
marginal-diagonal ordering). Table~\ref{tab:hankel-retention} reports
medians over layers; the balanced subspace dominates at every $r$, and
the gap is largest at small $r$.

\begin{table}[H]
\centering
\small
\setlength{\tabcolsep}{3pt}
\begin{tabular}{@{}lrrrrr@{}}
\toprule
Model & $r$ & Balanced & Hankel-diag & Saliency & Energy \\
\midrule
130M & 8  & 89.6 & 42.7 & 37.1 & 23.8 \\
130M & 16 & 93.3 & 47.4 & 54.3 & 41.2 \\
130M & 32 & 96.6 & 70.0 & 69.5 & 61.2 \\
370M & 8  & 86.9 & 32.3 & 33.4 & 26.1 \\
370M & 16 & 91.5 & 44.3 & 42.5 & 39.8 \\
370M & 32 & 95.5 & 65.7 & 63.9 & 55.7 \\
780M & 8  & 87.8 & 36.6 & 35.0 & 31.5 \\
780M & 16 & 92.3 & 51.5 & 52.1 & 45.8 \\
780M & 32 & 96.1 & 68.5 & 68.3 & 62.6 \\
1.3B & 8  & 86.8 & 29.0 & 28.0 & 18.6 \\
1.3B & 16 & 91.7 & 42.0 & 42.1 & 36.7 \\
1.3B & 32 & 96.0 & 67.0 & 66.3 & 58.1 \\
2.7B & 8  & 86.2 & 31.5 & 31.6 & 22.2 \\
2.7B & 16 & 91.6 & 44.8 & 46.5 & 38.2 \\
2.7B & 32 & 95.6 & 61.9 & 62.0 & 57.2 \\
\bottomrule
\end{tabular}
\caption{Fraction (\%) of the empirical Hankel spectral energy
$\operatorname{tr}(W_cW_o)$ retained by an $r$-dimensional subspace
(median over layers). Coordinate columns select native coordinates by the
layer-mean score $\mathbb{E}[H^2]$ (energy), $\mathbb{E}[H^2C^2]$
(saliency), or the strongest marginal-diagonal ordering
$\operatorname{diag}(W_c)\odot\operatorname{diag}(W_o)$.}
\label{tab:hankel-retention}
\end{table}

\section{Full-Test-Set Evaluation}
\label{app:fullppl}

Table~\ref{tab:fullppl} re-evaluates dense and \method{} on the
\emph{full} WikiText-2 test file (280 segments of 1{,}024 tokens,
286{,}440 tokens) instead of the 8/4-segment protocol of the main table,
with 95\% confidence intervals from a paired segment-level bootstrap
(10{,}000 resamples). Dense perplexity differs from the main table
(22.69/15.87/11.71 here vs.\ 25.21/17.97/12.39 on the first 8/4
segments) because the full file is a different token sample; all
comparisons are within-protocol. The full-set costs agree with the main
table for $r{=}64$ and are slightly smaller at $r{=}32$; in particular
the 1.3B $r{=}32$ headline ($+22.7\%$ on four segments) is
conservative---the full test set gives $+16.7\%$.

\begin{table}[H]
\centering
\small
\setlength{\tabcolsep}{3pt}
\begin{tabular}{@{}lcccc@{}}
\toprule
Model & dense ppl & $r{=}64$ & 95\% CI & $r{=}32$ \\
\midrule
130M & 22.69 & $+6.8\%$ & $[6.5, 7.2]$ & $+24.7\%$ \\
370M & 15.87 & $+5.9\%$ & $[5.6, 6.1]$ & $+20.2\%$ \\
1.3B & 11.71 & $+4.9\%$ & $[4.7, 5.1]$ & $+16.7\%$ \\
\bottomrule
\end{tabular}
\caption{Full-test-file evaluation (280$\times$1024 tokens): perplexity
of the uncompressed model and \method{} relative change; CIs are for the
$r{=}64$ change (the $r{=}32$ CIs are within $\pm1.3$ points,
[23.5, 26.0]/[19.4, 21.1]/[16.0, 17.4]).}
\label{tab:fullppl}
\end{table}

\section{State Quantization at Matched Budgets}
\label{app:quant}

Table~\ref{tab:quant} compares subspace reduction with per-mode
symmetric state quantization at matched budgets, using the same
token-by-token loop (rotation included for \method{} rows; integer rows
use per-mode absmax scales recomputed each step, without fused kernels
or clipping calibration). The best configuration depends on the budget
and scale. At 2$\times$, 16-bit quantization is close to lossless on
130M/370M but degrades 1.3B, leaving little room for truncation on the
smaller models. At 4$\times$, SPAR-R with int16 is best on 130M/370M,
while dense int8 remains best on 1.3B. At 8$\times$, dense int4
collapses, and only subspace plus int8 remains functional; SPAR-R leads
on 130M/370M, uniform \method{} on 1.3B. At 16$\times$, SPAR-R plus int8 is
best on all three scales. Quantization therefore dominates at mild
budgets, and subspace reduction extends the usable bit-width range at
aggressive ones (Table~\ref{tab:quant}).

\begin{table}[H]
\centering
\small
\setlength{\tabcolsep}{3pt}
\begin{tabular}{@{}lccc@{}}
\toprule
Configuration & 130M & 370M & 1.3B \\
\midrule
\multicolumn{4}{@{}l}{\textit{4096 bits/head (1$\times$)}} \\
Dense, $r{=}128$, fp32 & 25.21 & 17.97 & 12.39 \\
\addlinespace[2pt]
\multicolumn{4}{@{}l}{\textit{2048 bits/head (2$\times$)}} \\
Dense, $r{=}128$, int16 & \textbf{25.21} & \textbf{17.97} & 13.36 \\
BASS, $r{=}64$, fp32 & 27.02 & 19.00 & 13.01 \\
BASS+SPAR-R, $\bar r{=}64$, fp32 & 25.95 & 18.46 & \textbf{12.63} \\
\addlinespace[2pt]
\multicolumn{4}{@{}l}{\textit{1024 bits/head (4$\times$)}} \\
Dense, $r{=}128$, int8 & 28.39 & 18.79 & \textbf{12.69} \\
BASS, $r{=}64$, int16 & 27.03 & 19.00 & 13.88 \\
BASS+SPAR-R, $\bar r{=}64$, int16 & \textbf{25.93} & \textbf{18.45} & 13.62 \\
BASS, $r{=}32$, fp32 & 31.61 & 21.62 & 15.20 \\
BASS+SPAR-R, $\bar r{=}32$, fp32 & 30.20 & 19.67 & 13.41 \\
\addlinespace[2pt]
\multicolumn{4}{@{}l}{\textit{512 bits/head (8$\times$)}} \\
Dense, $r{=}128$, int4 & 217.2 & 98.9 & 34.36 \\
BASS, $r{=}64$, int8 & 30.80 & 20.00 & \textbf{13.40} \\
BASS+SPAR-R, $\bar r{=}64$, int8 & \textbf{29.75} & \textbf{19.49} & 14.04 \\
\addlinespace[2pt]
\multicolumn{4}{@{}l}{\textit{256 bits/head (16$\times$)}} \\
BASS, $r{=}32$, int8 & 35.37 & 22.50 & 15.38 \\
BASS+SPAR-R, $\bar r{=}32$, int8 & \textbf{34.49} & \textbf{20.79} & \textbf{14.77} \\
\bottomrule
\end{tabular}
\caption{WikiText-2 perplexity (lower is better) at matched per-head
state budgets (bits = retained modes $\times$ bits per value). $r$:
modes per layer; $\bar r$: average for SPAR-R, whose per-layer ranks vary;
compression is relative to the 4{,}096-bit dense state. Bold marks the
best entry for each model within a budget. All rows use the same
token-by-token loop; integer rows use per-mode absmax scales without
fused kernels or clipping calibration.}
\label{tab:quant}
\end{table}

\section{Perplexity versus Retained Rank}
\label{app:rank}

\begin{figure*}[t]
\centering
\includegraphics[width=0.88\linewidth]{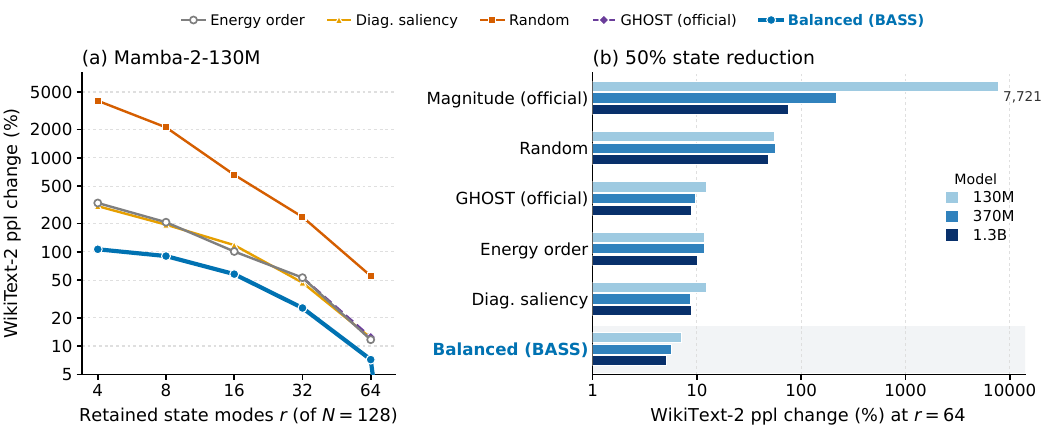}
\caption{(a) Perplexity change vs.\ retained rank for Mamba2-130M (log
scale). (b) At 50\% state reduction for the three checkpoints with complete
method sweeps (log scale); 780M/2.7B are in Table~\ref{tab:main}.}
\label{fig:ppl}
\end{figure*}

Figure~\ref{fig:ppl} gives the full rank sweep behind Table~\ref{tab:main}. The
ordering is stable across ranks: \method{} is lowest at every budget tested,
the diagonal criteria are next, and static magnitude is neither competitive nor
monotone in the budget.

% Generated by tools/make_appendix_length.py -- do not edit by hand.
\section{Context-Length Evaluation}
\label{app:length}

\noindent\textbf{Protocol.} The model is calibrated once, at $L{=}1024$, and
evaluated on WikiText-2 test text truncated to $L$ tokens; every method sees
the same segments, and all five checkpoints are evaluated.
Table~\ref{tab:length-abs} reports absolute perplexity at the two ends of the
sweep, where the dense model's own perplexity differs by an order of
magnitude, and Table~\ref{tab:length-pct} the cost relative to the dense
model at the same length.

\begin{table}[t]
\centering
\caption{Absolute perplexity at the shortest and longest evaluated context
at $50\%$ retention ($r{=}64$).
The compression error grows with $L$ in absolute terms far more slowly than
the dense perplexity falls, which is why the relative cost in
Table~\ref{tab:length-pct} rises.}
\label{tab:length-abs}
\footnotesize
\setlength{\tabcolsep}{4pt}
\begin{tabular}{@{}llcccc@{}}
\toprule
Model & $L$ & Dense & BASS & SPAR-R & GHOST \\
\midrule
130M & 16 & 195.18 & 199.52 & 198.33 & 202.21 \\
 & 2048 & 21.95 & 24.02 & 22.73 & 25.13 \\
\midrule
370M & 16 & 151.25 & 155.12 & 154.23 & 156.04 \\
 & 2048 & 15.90 & 16.96 & 16.34 & 17.61 \\
\midrule
780M & 16 & 128.64 & 129.25 & 128.69 & 133.42 \\
 & 2048 & 13.05 & 13.41 & 13.14 & 14.43 \\
\midrule
1.3B & 16 & 111.11 & 113.05 & 112.66 & 113.61 \\
 & 2048 & 10.79 & 11.47 & 11.04 & 11.89 \\
\midrule
2.7B & 16 & 104.36 & 104.43 & 104.02 & 106.29 \\
 & 2048 & 9.56 & 9.71 & 9.64 & 10.27 \\
\bottomrule
\end{tabular}
\end{table}

\begin{table*}[t]
\centering
\caption{Relative perplexity cost (\%) against the dense model at the same
context length, one calibration at $L{=}1024$. Within each model, rows are
grouped by budget: $r{=}64$ ($50\%$ retention), then $r{=}32$ ($25\%$).
Saliency is shown only at $r{=}64$.}
\label{tab:length-pct}
\footnotesize
\setlength{\tabcolsep}{3pt}
\renewcommand{\arraystretch}{0.96}
\begin{tabular}{@{}llcccccccc@{}}
\toprule
Model & Configuration & 16 & 32 & 64 & 128 & 256 & 512 & 1024 & 2048 \\
\midrule
130M & BASS r=64 & +2.2 & +3.0 & +3.5 & +3.8 & +4.2 & +5.2 & +7.2 & +9.4 \\
 & BASS+SPAR-R r=64 & +1.6 & +2.2 & +2.8 & +2.5 & +2.7 & +2.7 & +2.8 & +3.5 \\
 & Saliency r=64 & +3.8 & +5.6 & +6.8 & +7.6 & +8.5 & +9.4 & +12.2 & +14.6 \\
 & GHOST r=64 & +3.6 & +5.3 & +6.7 & +7.3 & +8.6 & +9.8 & +12.4 & +14.5 \\
\cmidrule(lr){1-10}
 & BASS r=32 & +4.6 & +6.2 & +8.2 & +10.3 & +13.1 & +17.8 & +25.4 & +33.4 \\
 & BASS+SPAR-R r=32 & +5.0 & +6.9 & +8.6 & +9.9 & +11.8 & +14.6 & +19.8 & +25.1 \\
 & GHOST r=32 & +8.7 & +14.6 & +21.1 & +27.2 & +33.9 & +41.8 & +53.4 & +64.7 \\
\midrule
370M & BASS r=64 & +2.6 & +2.6 & +3.3 & +3.8 & +4.2 & +4.3 & +5.7 & +6.6 \\
 & BASS+SPAR-R r=64 & +2.0 & +1.5 & +1.9 & +2.3 & +2.5 & +2.5 & +2.7 & +2.8 \\
 & Saliency r=64 & +2.9 & +3.7 & +4.6 & +5.3 & +6.1 & +7.2 & +8.7 & +10.0 \\
 & GHOST r=64 & +3.2 & +4.4 & +5.5 & +6.2 & +7.2 & +8.3 & +9.6 & +10.7 \\
\cmidrule(lr){1-10}
 & BASS r=32 & +5.7 & +6.2 & +8.4 & +10.3 & +11.9 & +13.9 & +20.3 & +26.0 \\
 & BASS+SPAR-R r=32 & +4.7 & +5.1 & +5.6 & +6.5 & +6.6 & +7.0 & +9.5 & +11.2 \\
 & GHOST r=32 & +8.2 & +26.8 & +140.6 & +737.4 & +1273.1 & +1939.5 & +2458.0 & +2412.4 \\
\midrule
780M & BASS r=64 & +0.5 & +0.7 & +1.2 & +1.0 & +1.1 & +1.1 & +1.7 & +2.8 \\
 & BASS+SPAR-R r=64 & +0.0 & +0.3 & +0.9 & +0.8 & +0.4 & +0.3 & +0.5 & +0.7 \\
 & Saliency r=64 & +4.3 & +5.6 & +6.1 & +6.4 & +6.8 & +6.9 & +8.0 & +9.1 \\
 & GHOST r=64 & +3.7 & +5.1 & +6.0 & +6.6 & +7.4 & +7.8 & +9.3 & +10.6 \\
\cmidrule(lr){1-10}
 & BASS r=32 & +2.2 & +2.9 & +3.9 & +4.2 & +5.0 & +5.9 & +9.0 & +13.0 \\
 & BASS+SPAR-R r=32 & +2.3 & +2.6 & +3.5 & +3.9 & +3.8 & +3.7 & +4.4 & +5.8 \\
 & GHOST r=32 & +9.4 & +16.6 & +44.7 & +118.3 & +204.6 & +304.6 & +584.5 & +880.1 \\
\midrule
1.3B & BASS r=64 & +1.8 & +1.6 & +2.3 & +2.7 & +2.8 & +2.5 & +5.1 & +6.2 \\
 & BASS+SPAR-R r=64 & +1.4 & +1.2 & +1.5 & +1.4 & +1.4 & +1.2 & +1.8 & +2.2 \\
 & Saliency r=64 & +3.2 & +4.5 & +4.5 & +5.1 & +5.5 & +5.8 & +8.9 & +10.5 \\
 & GHOST r=64 & +2.3 & +3.3 & +3.2 & +4.0 & +4.7 & +5.3 & +8.8 & +10.2 \\
\cmidrule(lr){1-10}
 & BASS r=32 & +4.1 & +4.7 & +7.2 & +9.6 & +10.7 & +12.0 & +22.8 & +28.0 \\
 & BASS+SPAR-R r=32 & +3.4 & +3.5 & +4.3 & +4.5 & +4.7 & +4.8 & +8.3 & +10.7 \\
 & GHOST r=32 & +6.8 & +10.7 & +14.3 & +24.7 & +36.9 & +54.4 & +83.7 & +156.3 \\
\midrule
2.7B & BASS r=64 & +0.1 & +0.3 & +0.3 & +0.5 & +0.5 & +0.7 & +1.3 & +1.5 \\
 & BASS+SPAR-R r=64 & -0.3 & +0.1 & +0.3 & +0.9 & +0.7 & +0.9 & +0.9 & +0.8 \\
 & Saliency r=64 & +2.3 & +4.5 & +4.6 & +4.5 & +4.2 & +4.1 & +5.0 & +5.7 \\
 & GHOST r=64 & +1.8 & +3.7 & +3.9 & +4.5 & +4.4 & +4.7 & +6.6 & +7.4 \\
\cmidrule(lr){1-10}
 & BASS r=32 & +1.2 & +1.7 & +2.1 & +2.5 & +2.9 & +3.4 & +6.5 & +8.1 \\
 & BASS+SPAR-R r=32 & +1.2 & +1.9 & +1.8 & +2.1 & +2.3 & +2.4 & +3.4 & +4.1 \\
 & GHOST r=32 & +5.5 & +10.4 & +13.2 & +16.6 & +21.7 & +25.1 & +33.2 & +38.2 \\
\bottomrule
\end{tabular}
\end{table*}

\end{document}